\documentclass{article}

\usepackage{array}

\usepackage{PRIMEarxiv}

\usepackage[utf8]{inputenc} 
\usepackage[T1]{fontenc}    
\usepackage{hyperref}       
\usepackage{url}            
\usepackage{booktabs}       
\usepackage{amsfonts}       
\usepackage{nicefrac}       
\usepackage{microtype}      
\usepackage{lipsum}
\usepackage{fancyhdr}       
\usepackage{graphicx}       
\graphicspath{{media/}}     

\title{LingShu: A Large-Scale Symptom-Centric Contextualized Knowledge Graph Bridging Traditional Chinese Medicine and Modern Biomedicine
}

\author{
  Rui Hua\textsuperscript{\textdagger,1},
  Zixin Shu\textsuperscript{\textdagger,2,3,4},
  Kai Chang\textsuperscript{1},
  Dengying Yan\textsuperscript{1},
  Jianan Xia\textsuperscript{1},
  Hui Zhu\textsuperscript{2,3,4} \\
  Shujie Song\textsuperscript{2,3,4},
  Shurui Yang\textsuperscript{5,6,7},
  Tongxin Wang\textsuperscript{8},
  Yue Yin\textsuperscript{9},
  Yu Wei\textsuperscript{10,11,12},
  Lijuan Pei\textsuperscript{10,11,12}, \\
  Yunhui Hu\textsuperscript{10,11,12},
  Hao Xu\textsuperscript{8},
  Mingzhong Xiao\textsuperscript{2,3,4},
  Xiaodong Li\textsuperscript{2,3,4},
  Haibin Yu\textsuperscript{13} \\
  Runshun Zhang\textsuperscript{14},
  Wenjia Wang\textsuperscript{*,10,11,12},
  Baoyan Liu\textsuperscript{*,15},
  Xuezhong Zhou\textsuperscript{*,1} \\
  \\
  \textsuperscript{1}Department of Computer Science and Technology, Beijing Jiaotong University, Beijing, China \\
  \textsuperscript{2}Institute of Liver Diseases, Hubei Key Laboratory of the theory and application research of \\  liver and kidney in traditional Chinese medicine, \\ Hubei Provincial Hospital of Traditional Chinese Medicine, Wuhan 430061, China \\
  \textsuperscript{3}Affiliated Hospital of Hubei University of Chinese Medicine, Wuhan 430061, China \\
  \textsuperscript{4}Hubei Province Academy of Traditional Chinese Medicine, Wuhan 430074, China \\
  \textsuperscript{5}Acupuncture and Moxibustion Department, Hubei Provincial Hospital of Traditional Chinese Medicine, \\ Affiliated Hospital of Hubei University of Chinese Medicine, Wuhan 430060, Hubei Province, China \\
  \textsuperscript{6}Hubei Provincial Clinical Research Center for Acupuncture and \\ Moxibustion in Obesity Treatment, Wuhan 430060, Hubei Province, China \\
  \textsuperscript{7}Hubei Shizhen Laboratory, Wuhan 430060, Hubei Province, China \\
  \textsuperscript{8}Cardiovascular Diseases Center, National Clinical Research Center for Chinese Medicine Cardiology, \\ Xiyuan Hospital, China Academy of Chinese Medical Sciences, Beijing 100091, China \\
  \textsuperscript{9}Institute of Chinese Materia Medica, China Academy of Chinese Medical Sciences \\
  \textsuperscript{10}Tianjin Tasly Digital Chinese Medicine Technology Co., Ltd., Tianjin 300410, China \\
  \textsuperscript{11}Tasly Biopharmaceuticals Co., Ltd., Tianjin 300410, China \\
  \textsuperscript{12}State Key Laboratory of Chinese Medicine Modernization, Tianjin 300193, China \\
  \textsuperscript{13}The First Affiliated Hospital, Henan University of Chinese Medicine, Zhengzhou 450000, China \\
  \textsuperscript{14}Department of Gastroenterology, Guang'anmen Hospital,\\  China Academy of Chinese Medical Sciences, Beijing, China \\
  \textsuperscript{15}China Academy of Chinese Medical Sciences, Beijing 100700, China \\
  \\
  \textsuperscript{\textdagger}These authors contributed equally. \\
  \textsuperscript{*}Corresponding authors: Wenjia Wang, Baoyan Liu, and Xuezhong Zhou.
}

\begin{document}
\maketitle

\begin{abstract}

Biomedical knowledge graphs (KGs) are pivotal for knowledge organization, yet traditional binary relations often struggle to represent the conditional nature of biomedical knowledge.
Symptoms provide a shared phenotypic layer for linking Traditional Chinese Medicine (TCM), which relies on symptom patterns for syndrome differentiation and treatment selection, with modern biomedicine, which connects clinical manifestations to diseases and molecular mechanisms.
We present LingShu, a large-scale symptom-centric contextualized knowledge graph designed to bridge TCM and modern biomedicine. 
The exported version of LingShu analyzed in this study comprises 17.33 million atom-level entity records and 39.47 million relation records, including 17.19 million semantic triples and 22.29 million contextualized quadruples.
LingShu integrates multi-source data, including clinical electronic medical records, authoritative TCM texts, biomedical ontologies, and curated knowledge bases, through a pipeline combining natural language processing, terminology normalization, and human-in-the-loop verification. 
A key innovation of LingShu is its hybrid data model: it maintains 64 typed triple relation patterns to ensure broad connectivity, while incorporating 35 contextual quadruple relation patterns to capture conditional medical associations.
This dual-structure approach explicitly encodes conditional knowledge, providing a granular representation of the contexts associated with medical relations. 
These contextualized relations cover syndrome-dependent herb efficacy, disease-contextualized drug effects, population-specific clinical associations, and mechanism-related therapeutic responses. 
Furthermore, we developed a web platform (\url{http://www.tcmkg.com/}) that integrates graph visualization, graph-based reasoning, and an evidence-grounded knowledge question-answering agent.
By anchoring large language models to structured graph evidence, the system supports traceable, context-aware knowledge exploration for decision-support analysis and biological hypothesis generation.

\end{abstract}

\section{Introduction}

Knowledge graphs (KGs) represent entities and their interrelationships in a structured graphical format, offering an efficient paradigm for the integration, retrieval, and analysis of complex data \cite{fensel2020introduction}.
In the biomedical domain, rapid advancements in big data analytics and natural language processing have catalyzed the accumulation of extensive knowledge bases, ranging from gene--protein interaction networks to Traditional Chinese Medicine (TCM) graph and clinical electronic medical record (EMR) repositories \cite{yu2017knowledge,zheng2020tcmkg}.
Prominent examples of such resources include molecular biology databases such as UniProt \cite{uniprot2019uniprot} and Gene Ontology \cite{gene2019gene}, specialized TCM knowledge bases such as SymMap \cite{wu2019symmap} or TCMSP \cite{ru2014tcmsp}, and comprehensive clinical ontologies like UMLS \cite{bodenreider2004unified} and SNOMED CT \cite{donnelly2006snomed,schulz2023snomed}.
These resources lay a solid foundation for intelligent medical applications, including disease risk assessment, clinical diagnostic support, medical quality management, and knowledge-based question answering systems \cite{li2020real,gong2021smr}.

Symptom science represents a foundational component of biomedicine, bridging disease prevention, diagnosis, and treatment \cite{ashley2016towards}.
TCM prioritizes the holistic interconnection among symptoms, employing “syndrome differentiation” to guide individualized treatment \cite{packer2024traditional,wang2025tcmeval,wang2025rigorous}.
Conversely, modern biomedicine advances diagnostic and therapeutic innovations by deciphering symptom patterns indicative of underlying pathologies \cite{lowe2024persistent,subramanian2022symptoms,shin2023associations}.
In the context of TCM, symptoms are regarded as the primary phenotypic manifestations of internal physiological disharmony.
Rather than isolated indicators, they form intricate clusters that define specific "Syndromes" (Zheng), serving as the essential clinical evidence for determining therapeutic principles and herbal selection.
Integrating the empirical wisdom of TCM with the precise biological evidence of modern biomedicine offers valuable insights for optimizing symptom management and therapeutic strategies \cite{wang2025rigorous,cheang2024traditional,yang2023traditional,ji2024jinlida}.

Despite the growing recognition of symptom science, three challenges remain unresolved. First, existing medical KGs and ontologies often treat symptoms as subsidiary phenotypes of diseases rather than as central entities \cite{chandak2023building,bang2023biomedical,lu2023sympgan}. Second, TCM and modern biomedical resources use heterogeneous vocabularies and conceptual systems, making it difficult to align symptoms, syndromes, diseases, interventions, and molecular mechanisms. Third, conventional triple-based KGs cannot preserve the conditional structure of medical knowledge, such as syndrome-dependent treatment, gene-mediated drug effects, population-specific co-occurrence, and prescription modification. These limitations highlight the need for a symptom-centric KG that bridges the semantic, mechanistic, and contextual gaps between TCM and modern biomedicine.

Contextual knowledge representation has long been recognized as a fundamental challenge in artificial intelligence (AI)~\cite{guha1992contexts,mccarthy1997formalizing,polanyi1958personal}.
Since McCarthy first introduced the formalization of context \cite{mccarthy1994formalizing,mccarthy1997formalizing}, researchers have emphasized that knowledge is rarely universally true but holds within specific boundaries \cite{guha1992contexts}.
In the biomedical domain, understanding context is critical for analyzing complex clinical relationships.
Biomedical knowledge is inherently conditional. 
For instance, therapeutic relationships (e.g., drug--disease or drug--symptom associations) are often valid only under specific disease stages, genetic conditions, population groups, or pharmacological functional states~\cite{ashley2016towards}.
Indeed, many drugs exhibit multi-functional properties where their efficacy and safety profiles shift significantly according to the physiological context and diverse patient attributes \cite{lauschke2024pharmacogenomics}.
However, traditional knowledge graphs predominantly rely on "Subject-Predicate-Object" triples \cite{iglesias2023comparison,vrandevcic2014wikidata}.
This flat structure fails to capture the multidimensional nature of biomedical knowledge (e.g., \textit{Drug A treats Symptom B \textbf{given} Condition C}), limiting the precision of downstream reasoning and clinical applications \cite{iglesias2023comparison,vrandevcic2014wikidata}.

\begin{figure*}
    \centering
    \includegraphics[width=0.9\textwidth]{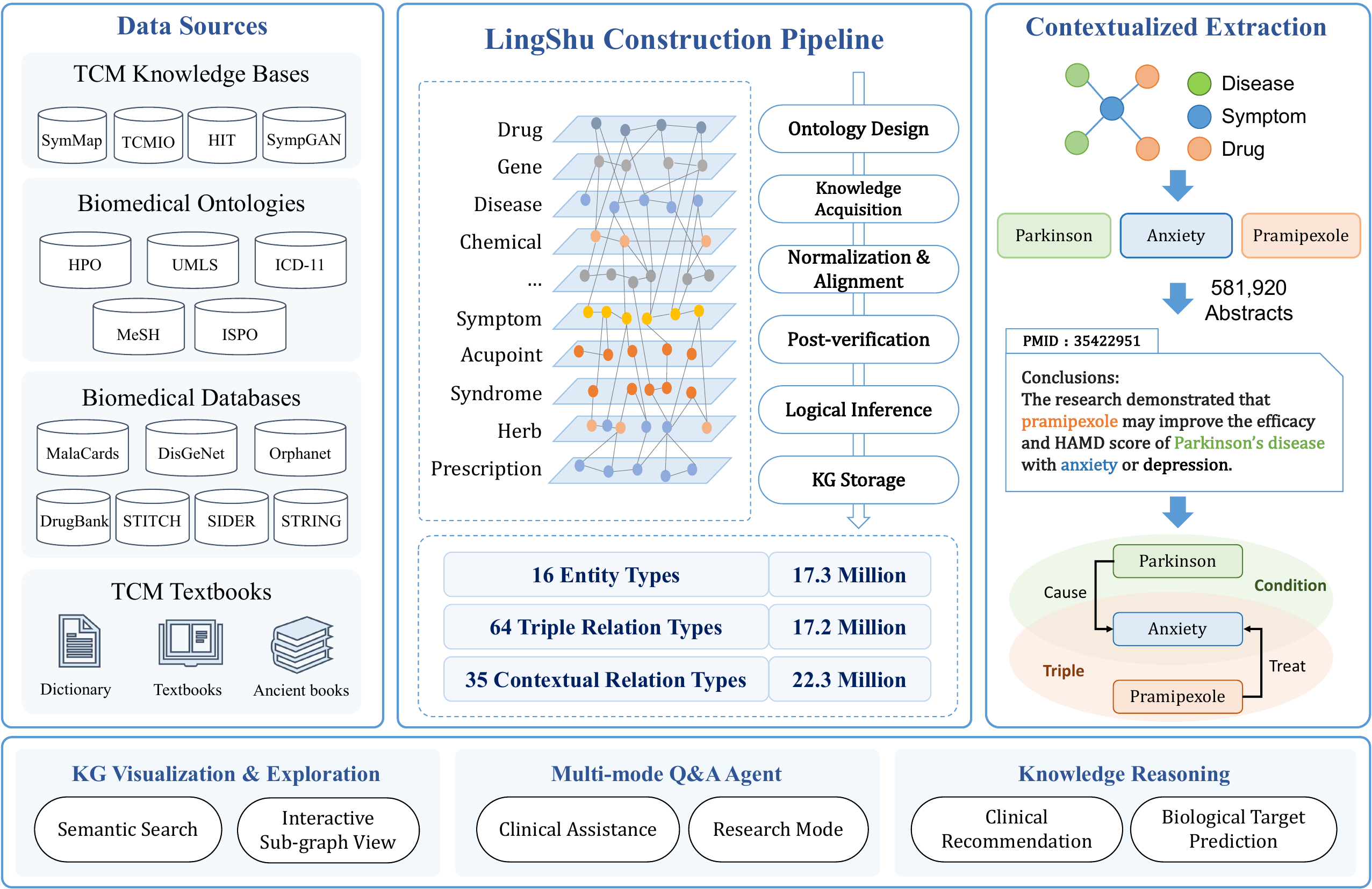}
    \caption{Overview of the LingShu framework. This framework integrates multi-source data through a standardized pipeline. The right panel illustrates the formalization of clinical conclusions into hyper-relational quadruples, such as treating Anxiety with Pramipexole under the specific condition of Parkinson’s disease. The bottom layer presents the integrated application system supported by LingShu.}
    \label{fig:overview}
\end{figure*}

To address these limitations, this study introduces a hyper-relational model to construct a high-quality, contextualized symptom knowledge graph, named LingShu. 
Instead of relying on simple triples, LingShu explicitly encodes conditional variables (such as disease contexts and molecular targets) using quadruples, thereby achieving a more expressive representation of biomedical facts.
The development of LingShu encompasses a comprehensive pipeline, including the integration of diverse data from clinical records and TCM literature, rigorous knowledge extraction, and terminology normalization, as illustrated in Figure \ref{fig:overview}.

The contributions of this work are as follows:

(1) We constructed LingShu, a large-scale symptom-centric KG that integrates 17.33 million atom-level entity records and bridges TCM with modern biomedicine by prioritizing symptom phenotypes as core entities.

(2) We proposed a contextualized schema that defines 35 typed contextual quadruple patterns using hyper-relational quadruples, enabling context-aware graph retrieval and reasoning beyond conventional triples.

(3) We developed a web-based platform that supports graph exploration, interpretable context-aware reasoning, multi-mode knowledge question-answering (Q\&A), and human-in-the-loop data management for continuous knowledge curation.

\section{Related Work}

Modern biomedical knowledge resources provide essential infrastructure for terminology standardization, knowledge integration, and computational reasoning. 
For LingShu, the most relevant lines of work include biomedical terminologies and literature resources, clinical phenotype resources, drug and molecular knowledge bases, TCM knowledge bases, and contextual knowledge representation. 
Together, these resources establish the foundation for biomedical knowledge graph construction, but they remain limited in symptom-centric cross-paradigm integration and conditional relation representation.

\subsection{Biomedical Terminologies and Literature Resources}

Large-scale biomedical terminologies and literature resources have been widely used to support semantic interoperability and information extraction. UMLS \cite{bodenreider2004unified} integrates multiple biomedical vocabularies and provides a unified semantic framework for concept normalization. MeSH \cite{lipscomb2000medical} supports controlled indexing of biomedical literature, while PubMed \cite{canese2013pubmed} provides a major evidence source for biomedical text mining and relation extraction. Other structured resources, such as Gene Ontology \cite{gene2019gene} and UniProt \cite{uniprot2019uniprot}, provide standardized descriptions of genes, proteins, biological functions, and molecular annotations.
These resources are indispensable for biomedical entity alignment and evidence retrieval. However, they are not primarily designed to represent TCM concepts, syndrome differentiation, herb compatibility, or symptom-pattern reasoning. As a result, direct integration of these resources is insufficient for building a symptom-centric graph that links TCM clinical knowledge with modern biomedical mechanisms.

\subsection{Clinical Phenotype Knowledge Bases}

Clinical phenotype resources provide structured representations of patient manifestations and disease-associated abnormalities. The Symptom Ontology \cite{noy2009bioportal} explicitly describes symptom concepts and their hierarchical organization. HPO \cite{kohler2021human} has become a widely used standard for phenotypic abnormalities, especially in genetic diseases. Disease-oriented resources such as MalaCards \cite{rappaport2017malacards}, Orphanet \cite{weinreich2008orphanet}, ICD-11 \cite{harrison2021icd}, and SNOMED CT \cite{donnelly2006snomed,schulz2023snomed} also include symptom and phenotype information as part of broader clinical terminology systems.
These resources have greatly advanced phenotype standardization and disease annotation. Nevertheless, most of them organize symptoms as attributes, manifestations, or annotations of diseases. This disease-centred organization limits the independent analysis of symptom phenotypes, cross-disease symptom patterns, and symptom-based reasoning. The limitation is particularly important for TCM, where symptoms and symptom clusters are central evidence for syndrome differentiation and treatment selection.

\subsection{Modern Biomedical Knowledge Graphs}

Modern biomedical KGs provide large-scale mechanistic and pharmacological representations for connecting diseases, drugs, genes, proteins, chemicals, pathways, phenotypes, and clinical observations.
Early integrative resources such as Hetionet systematically combined heterogeneous biomedical databases into a multi-relational network for disease biology and drug repurposing \cite{himmelstein2017systematic}.
More recent precision-medicine-oriented graphs have further expanded this paradigm.
PrimeKG integrates multi-scale biomedical knowledge to provide a disease-centered view spanning drugs, proteins, pathways, biological processes, phenotypes, and anatomical information \cite{chandak2023building}.
SPOKE constructs a large-scale biomedical knowledge engine by integrating dozens of biomedical databases into a semantically organized graph covering diverse biological and clinical entity types \cite{morris2023scalable}.
The clinical KG further demonstrates how KGs can support the interpretation of clinical proteomics data by integrating experimental measurements, public databases, and biomedical literature~\cite{santos2022knowledge}.
Together with resources such as DrugBank \cite{wishart2018drugbank}, SIDER \cite{kuhn2016sider}, STITCH \cite{kuhn2014stitch}, STRING \cite{szklarczyk2019string}, and DisGeNET \cite{pinero2020disgenet}, these KGs and databases provide essential infrastructure for mechanism-oriented analysis, drug repurposing, adverse-effect mining, and molecular association discovery.
Despite their scale and utility, existing modern biomedical KGs are primarily organized around diseases, drugs, and genes, and they rarely represent TCM-specific concepts such as syndromes, herbal prescription, acupoints, meridians, or herb addition and removal.
Moreover, symptom phenotypes are seldom used as the central organizing layer, and most resources still rely on binary relations, making it difficult to capture the conditional structure of medical knowledge across TCM and modern biomedicine.

\subsection{Traditional Chinese Medicine Knowledge Bases}

Efforts to formalize and digitize TCM knowledge have progressed from terminology standardization and ontology construction to large-scale databases and KG resources.
Early ontology-driven studies laid the foundation for structured TCM knowledge representation.
For example, UTCMLS~\cite{zhou2004ontology} introduced an ontology-based framework for organizing TCM terminology and supporting knowledge storage, concept-based retrieval, and information integration.
More recently, ISPO provided an integrated ontology of symptom phenotypes for semantic integration of TCM data, emphasizing the importance of symptom standardization for cross-source knowledge organization~\cite{shu2024ispo}.
In parallel, pharmacology-oriented TCM databases such as TCMID~\cite{huang2018tcmid}, TCMSP~\cite{ru2014tcmsp}, SymMap~\cite{wu2019symmap}, HERB~\cite{fang2021herb}, TCMIO~\cite{liu2020tcmio}, and HIT/HIT 2.0~\cite{ye2010hit,yan2022hit} have organized herbs, prescriptions, ingredients, targets, diseases, symptoms, and pharmacological evidence from different perspectives.
Specialized resources such as SoFDA~\cite{zhang2022sofda} and SympGAN~\cite{lu2023sympgan} further demonstrate the value of structured association resources for disease--syndrome--formula analysis and symptom--gene association networks.
Together, these resources provide important foundations for network pharmacology, target prediction, mechanism discovery, and symptom-related knowledge integration.

TCM KG studies have further used ontology design, named entity recognition, relation extraction, and graph construction methods to structure knowledge from classical texts, clinical records, medical cases, and domain databases~\cite{yu2017knowledge,zheng2020tcmkg,zhou2019research}.
Recent work has begun to explore the use of large language models (LLMs) in TCM KG construction and case-based question-answering applications~\cite{zhang2024traditional,duan2025research}.
Broader surveys of TCM KGs and AI in TCM further indicate the growing importance of structured, computable, and reasoning-oriented representations of TCM knowledge~\cite{qu2024review,yan2025artificial}.

Despite these advances, existing TCM resources remain fragmented across herbs, formulas, symptoms, syndromes, diseases, ingredients, and targets, and many emphasize pharmacological associations or target prediction rather than symptom-centric organization and syndrome-differentiated clinical knowledge.
In particular, prescription modification, population-specific clinical associations, and evidence provenance are still insufficiently represented within a unified graph schema, leaving conditional TCM knowledge difficult to model systematically.

\subsection{Contextual Knowledge Representation}



The importance of context in knowledge has long been recognized in philosophy and knowledge theory. 
In \textit{Personal Knowledge}, Polanyi argued that knowledge is not purely detached or context-free, but is shaped by tacit understanding, personal judgment, and the situation in which it is applied \cite{polanyi1958personal}. 
This perspective is particularly relevant to medicine, where the validity of a statement often depends on clinical background, patient characteristics, disease stage, therapeutic purpose, or interpretive framework. 
In artificial intelligence, context was later formalized as a central issue in knowledge representation, with McCarthy and subsequent studies emphasizing that many propositions are valid only under specific assumptions or situations~\cite{guha1992contexts,mccarthy1994formalizing,mccarthy1997formalizing}. 
In semantic web and KG research, several approaches have been proposed to represent contextualized or n-ary relations, including named graphs, reification, contextualized triples, qualified statements, and n-ary relation modeling \cite{alexander2006rdf,giunti2021representing,serafini2012contextualized,vrandevcic2014wikidata}. 
These approaches show that triples are often insufficient when a relation requires provenance, temporal scope, confidence, condition, or other qualifiers. Biomedical knowledge is especially dependent on such contextual constraints. 
Therapeutic effects, adverse reactions, clinical co-occurrence patterns, and mechanism-mediated relations may depend on disease stage, syndrome type, genetic background, molecular target, population group, or treatment composition. 
Prior studies have begun to address conditional biomedical statements and scientific KG representations involving conditions \cite{jiang2019role,jiang2020biomedical,xu2023conditional}. 
However, existing contextual representation frameworks have not been systematically adapted to integrate the contextual logic of TCM with modern biomedical mechanisms. 
In TCM, therapeutic knowledge is often conditioned by syndrome differentiation, formula compatibility, and prescription modification; in modern biomedicine, therapeutic and adverse relations may be constrained by disease contexts, molecular targets, genetic background, or patient populations. 
A symptom-centric KG that bridges TCM and modern biomedicine therefore requires not only entity alignment, but also a representation model capable of preserving the conditions under which medical relations hold. 
This motivates the contextualized quadruple schema used in LingShu.

\subsection{Limitations of Existing Works}

Existing resources collectively provide rich biomedical, pharmacological, phenotypic, and TCM knowledge. However, three gaps remain. First, symptom knowledge is distributed across disease-centred clinical resources, TCM databases, and molecular databases, with limited support for symptom-centric integration. Second, TCM and modern biomedical resources use different conceptual systems, making it difficult to align symptom patterns, syndromes, interventions, and molecular mechanisms. Third, conventional triples cannot adequately represent conditional medical knowledge, including syndrome-dependent treatment, disease-specific drug effects, gene-mediated mechanisms, population-specific co-occurrence, and prescription modification.

LingShu is designed to address these gaps by constructing a symptom-centric knowledge graph that bridges TCM and modern biomedicine. It integrates heterogeneous resources through a unified atom--concept representation, defines typed triple and contextual quadruple schemas, and uses hyper-relational quadruples to encode the context under which medical relations are valid.

\section{Materials and Methods}

This section describes the construction of LingShu as a reproducible knowledge acquisition and graph construction pipeline. We first define the symptom-centric ontology, including entity types, typed triple schemas, and contextual quadruple schemas. We then describe the heterogeneous data sources and the preprocessing workflow, including atom--concept normalization, source-specific knowledge extraction, entity alignment, post-verification, logical inference, and graph storage.

\subsection{Ontology Design}

\begin{figure*}
  \centering
  \includegraphics[width=0.9\textwidth]{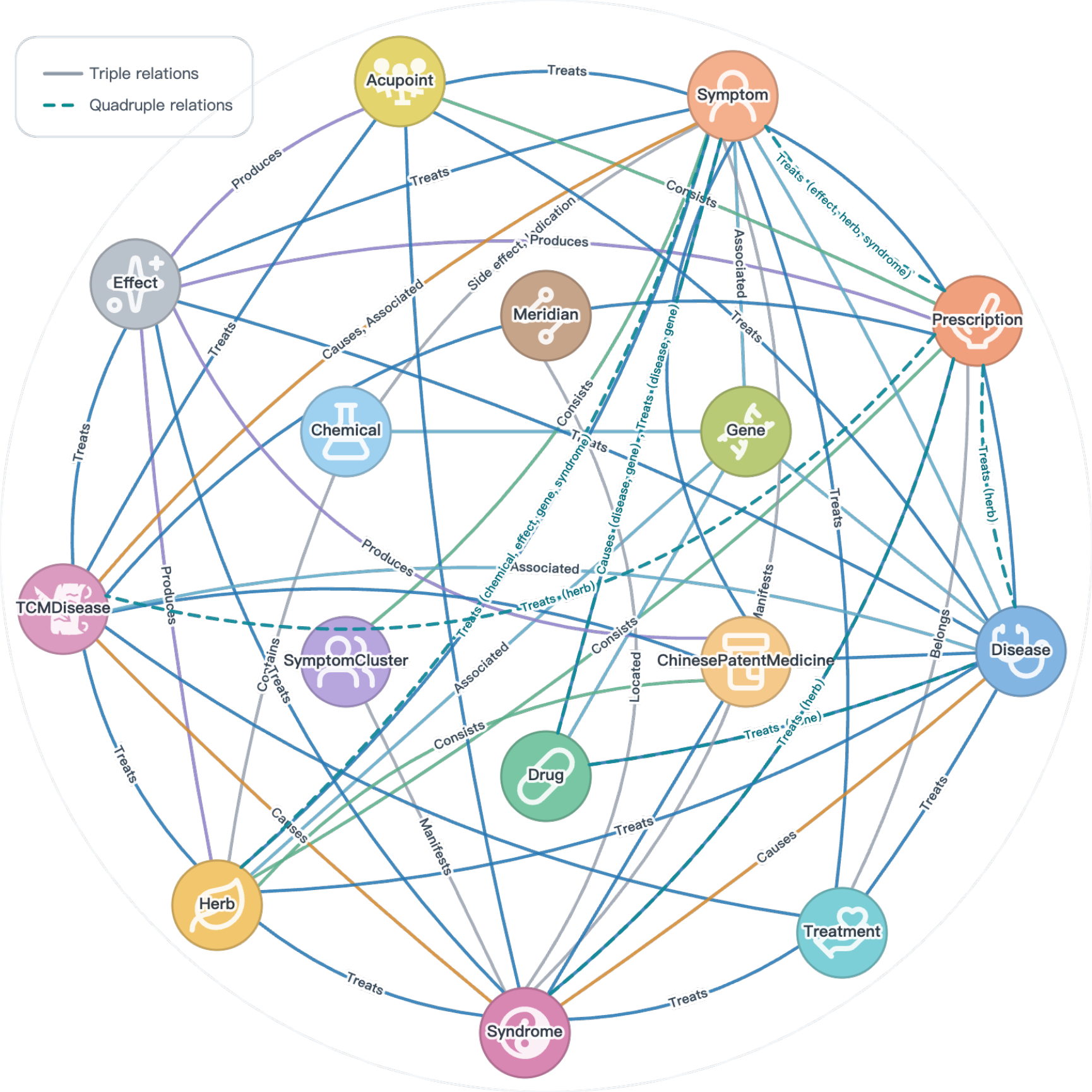}
  \caption{The ontology schema of LingShu.}
  \label{fig:schema}
\end{figure*}

To effectively bridge the semantic gap between the holistic concepts of TCM and the molecular mechanisms of modern biomedicine, we designed a rigorous schema centered on symptom phenotypes.
This schema defines the standardized vocabulary and logical structure for entities, basic predicates, typed relation patterns, and contextualized relations within LingShu.
The comprehensive ontology, which includes all defined entity types, relation categories, and hyper-relational quadruple edges, is shown in Figure \ref{fig:schema}.

\subsubsection{Entity Types}

LingShu encompasses 16 entity types spanning TCM and modern biomedicine, as follows:

(1) Symptom: A symptom is a subjective perception of discomfort or an objectively observable functional abnormality. Typical attributes include location, quality, severity, frequency, timing, and triggers.

(2) Symptom Cluster: A symptom cluster is a group of symptoms that commonly co-occur and are interrelated. In modern biomedicine, such clusters often indicate a specific medical condition (for example, metabolic syndrome). In TCM, symptom clusters reflect a particular syndrome (pattern), such as Liver Qi stagnation, which may include depressed mood, chest oppression, and a bitter taste in the mouth.

(3) Herb: Herbs are natural medicinal materials used under TCM theory for disease prevention, treatment, rehabilitation, and health maintenance. Attributes include the Four Natures, Five Flavors, channel tropism, and toxicity. For example, ginseng is sweet, slightly bitter, and slightly warm; it enters the Spleen and Lung Channels.

(4) Prescription: A prescription is a formulation created by combining two or more herbs according to principles based on etiology, pathogenesis, and clinical presentation. For example, Four-Gentlemen Decoction consists of ginseng, atractylodes, poria, and honey-fried licorice root.

(5) Chinese Patent Medicine: Chinese Patent Medicine refers to standardized finished TCM medicinal products, usually produced according to fixed formulations, dosage forms, and approved indications. It complements prescription entities by representing marketed or prepared TCM products rather than source-text formula descriptions.

(6) Syndrome: In TCM, a syndrome is a specific pathological state composed of interrelated symptoms and signs, forming the basis for diagnosis and treatment. For example, the Qi Deficiency pattern features shortness of breath, fatigue, mental lassitude, and reluctance to speak.

(7) Therapeutic Efficacy: Therapeutic Efficacy refers to the concrete therapeutic actions and outcomes exhibited by an herb or prescription in disease prevention and treatment.
Each herb or formula, determined by its constituents and compatibility principles, has characteristic therapeutic efficacies. 
For example, Four-Gentlemen Decoction is a tonifying formula used for supplementation, with the therapeutic efficacy of tonifying Qi and strengthening the Spleen.

(8) Therapeutic Method: Therapeutic Method denotes the therapeutic principle or clinical intervention strategy in TCM, such as clearing heat, resolving phlegm, or tonifying qi. Prescriptions are linked to therapeutic methods to encode formula-indication logic.

(9) Disease: A disease is an abnormal state or dysfunction arising from specific pathological processes, which may lead to impaired physiological functions and a range of clinical manifestations.

(10) TCM Disease: TCM Disease refers to disease names defined within the TCM nomenclature system, such as cough or consumptive disease. It complements modern disease entities and supports cross-paradigm alignment of clinical concepts.

(11) Drug: Drugs are medicines obtained via chemical synthesis or natural extraction. They are typically developed and administered as single active ingredients to achieve specific therapeutic effects. For example, aspirin’s principal component is salicylic acid, which has antipyretic, analgesic, and anti-inflammatory effects.

(12) Chemical: A chemical is a chemically pure substance with a defined molecular structure, which may serve as a raw material for drugs or exhibit distinct pharmacological activities. For example, salicylic acid is a principal component in aspirin and is widely used in other pharmaceutical products.
  
(13) Gene/Protein: Gene/Protein entities represent molecular targets and functional units used to connect clinical phenotypes, drugs, chemicals, herbs, and diseases. In LingShu, genes and proteins are represented under the unified Gene entity type and aligned using standard gene identifiers.

(14) Meridian: Meridian denotes the channel network in TCM theory through which qi and blood circulate, serving as anatomical context for syndrome localization.

(15) Acupoint: Acupoint refers to specific points on meridians used in acupuncture and related therapies, linked to symptoms, syndromes, and therapeutic efficacies.

(16) Population: Population entities denote disease-specific patient cohorts used to encode population-specific clinical co-occurrence contexts in LingShu.

\subsubsection{Triple Relations}

LingShu draws on semantic types from resources such as the Unified Medical Language System (UMLS) \cite{bodenreider2004unified} and defines relation patterns as typed schemas rather than relations alone.
A triple relation pattern is represented as $(h_{type}, r, t_{type})$, where $h_{type}$ and $t_{type}$ denote the entity types of the head and tail entities, and $r$ denotes the basic semantic predicate.
This design separates predicate-level semantics from entity-type constraints: the predicate specifies the semantic meaning of a relation, whereas the head and tail entity types define its admissible domain and range.
The definitions of the 13 basic relations are summarized in Table~\ref{tab:relation_definitions}.

\begin{table*}[ht]
  \centering
  \caption{Definitions of the 13 basic relations used in LingShu. These relations specify the semantic meaning of triples, while admissible head and tail entity types are constrained separately by the typed triple patterns in the ontology.}
  \label{tab:relation_definitions}
  \renewcommand{\arraystretch}{1.12}
  \begin{tabular}{p{0.12\textwidth}p{0.84\textwidth}}
  \toprule
  \textbf{Relation} & \textbf{Definition} \\
  \midrule
  \textit{treats} & Indicates a therapeutic relation in which a prescription, herb, Chinese patent medicine, acupoint, drug, therapeutic efficacy, or therapeutic method is reported to alleviate, manage, or treat a symptom, syndrome, TCM disease, modern disease, or related pathological state. \\
  \textit{causes} & Indicates a causal, contributing, triggering, or adverse relation in which one concept is reported to induce, aggravate, increase the risk of, or contribute to another clinical condition, phenotype, or disease state. \\
  \textit{produces} & Indicates that a concrete TCM entity, such as an herb, prescription, Chinese patent medicine, or acupoint, gives rise to a therapeutic action or functional effect. \\
  \textit{consists of} & Indicates a part--whole relation in which a complex concept is composed of one or more constituent concepts, such as a prescription or Chinese patent medicine composed of herbs or acupoints, or a symptom cluster composed of symptoms. \\
  \textit{contains} & Indicates a containment relation in which an herb includes a chemical or biologically active component. \\
  \textit{manifests as} & Indicates a manifestation relation in which a TCM syndrome is characterized by observable symptoms or symptom clusters. \\
  \textit{associated with} & Indicates a non-directional association supported by biomedical evidence, statistical co-occurrence, or curated database knowledge. \\
  \textit{interacts with} & Indicates a reciprocal interaction between biomedical concepts, including molecular, chemical, or pharmacological interactions. \\
  \textit{has side effect} & Indicates an adverse-effect association in which a chemical or biomedical exposure is linked to a reported symptom or clinical phenotype. \\
  \textit{has indication} & Indicates an indication-related association in which a chemical or biomedical concept is linked to a symptom or clinical phenotype for which it is reported to be clinically relevant. \\
  \textit{belongs to} & Indicates a membership or categorization relation in which a prescription or Chinese patent medicine is assigned to a broader therapeutic category or therapeutic method. \\
  \textit{located in} & Indicates a TCM localization relation that assigns a syndrome or clinical pattern to a meridian-related location. \\
  \textit{is a} & Indicates a hierarchical subsumption relation between concepts, where the head concept is a more specific subtype or narrower concept of the tail concept. \\
  \bottomrule
  \end{tabular}
\end{table*}

Under this definition, LingShu contains 64 typed triple relation patterns instantiated from 13 basic relations: \textit{Treats}, \textit{Causes}, \textit{Produces}, \textit{ConsistOf}, \textit{Contains}, \textit{ManifestationOf}, \textit{AssociatedWith}, \textit{InteractsWith}, \textit{SideEffect}, \textit{Indication}, \textit{BelongsTo}, \textit{LocatedIn}, and \textit{isa}.
These relations cover the major semantic structures required for integrating TCM and modern biomedical knowledge.
For example, \textit{Treats} links TCM therapeutic entities, such as herbs, prescriptions, Chinese patent medicines, therapeutic methods, therapeutic efficacies, and acupoints, to symptoms, syndromes, TCM disease names, or diseases, while also covering modern drug--symptom and drug--disease therapeutic relations.
\textit{ConsistOf} and \textit{Contains} encode compositional knowledge, such as prescriptions or Chinese patent medicines composed of herbs or herbs containing chemical components.
\textit{Produces} links herbs, prescriptions, Chinese patent medicines, or acupoints to therapeutic efficacies.
\textit{ManifestationOf} connects syndromes with their manifested symptoms or symptom clusters.
\textit{Causes} represents etiological or adverse associations involving drugs, herbs, symptoms, or syndromes.
\textit{AssociatedWith} and \textit{InteractsWith} capture statistical, molecular, pharmacological, or mechanistic associations among biomedical entities.
The remaining relations encode pharmacological evidence, hierarchical structure, therapeutic-method membership, and TCM-specific localization. 

All typed relation patterns were predefined in the ontology and used as hard constraints during information extraction, entity alignment, relation validation, and graph construction.

\subsubsection{Contextual Quadruple Relations}

In conventional knowledge graph construction, medical knowledge is usually represented as triples in the form $(h, r, t)$, where $h$ and $t$ denote the head and tail entities, and $r$ denotes the semantic relation between them.
However, medical knowledge is often context-dependent: therapeutic efficacy, adverse effects, mechanistic associations, and clinical co-occurrence patterns may hold only under specific syndromic, disease, molecular, compositional, or population contexts.
To capture this conditional nature, LingShu adopts a hyper-relational representation by introducing an explicit contextual variable.
A contextual knowledge unit is formally represented as a quadruple $q=(h, r, t, c)$, where $c$ denotes the condition or context that constrains the validity of the relation $(h, r, t)$.

Similar to triple relations, contextual relations in LingShu are defined as typed schemas rather than relation labels alone.
A contextual quadruple pattern is represented as $(h_{type}, r, t_{type}, c_{type})$, where $h_{type}$, $t_{type}$, and $c_{type}$ denote the entity types of the head, tail, and contextual entities, respectively.
Under this definition, LingShu defines 35 typed contextual quadruple patterns.
These patterns are organized around three basic contextual semantics: condition-aware treatment, condition-aware causation, and condition-aware co-occurrence.
The contextual conditions and their semantic definitions are summarized in Table~\ref{tab:condition_definitions}.

\begin{table*}[ht]
  \centering
  \caption{Definitions of contextual conditions used in LingShu. Each contextualized quadruple is represented as $(h,r,t,c)$, where $c$ specifies the condition under which the relation $(h,r,t)$ holds.}
  \label{tab:condition_definitions}
  \renewcommand{\arraystretch}{1.12}
  \begin{tabular}{p{0.17\textwidth}p{0.78\textwidth}}
  \toprule
  \textbf{Condition} & \textbf{Definition} \\
  \midrule
  \textit{Constituent herb} & Indicates that a prescription treats a symptom through one of its constituent herbs. \\
  \textit{Added herb} & Indicates that a prescription treats a symptom, syndrome, TCM disease, or modern disease after a specific herb is added to the prescription. \\
  \textit{Removed herb} & Indicates that a prescription treats a symptom, syndrome, TCM disease, or modern disease after a specific herb is removed from the prescription. \\
  \textit{Syndrome} & Indicates that a therapeutic relation is valid under a specific TCM syndrome context, reflecting syndrome differentiation in TCM. \\
  \textit{Gene} & Indicates that a therapeutic or adverse relation is mediated by, associated with, or conditioned on a gene-related mechanism. \\
  \textit{Population} & Indicates that a co-occurrence relation is statistically supported within a specific disease-defined population cohort. \\
  \textit{Disease} & Indicates that a therapeutic or adverse relation is valid under a specific disease context. \\
  \textit{Chemical} & Indicates that a therapeutic relation is mediated by, associated with, or conditioned on a chemical component. \\
  \textit{Therapeutic efficacy} & Indicates that a therapeutic relation is mediated by or associated with a specific therapeutic efficacy. \\
  \bottomrule
  \end{tabular}
\end{table*}

Condition-aware treatment represents therapeutic relations whose validity depends on an explicit context.
For example, an herb or prescription may treat a symptom only under a specific TCM syndrome; a drug may treat a symptom under a disease context; and a drug, herb, or prescription may exert therapeutic effects through a gene, chemical component, therapeutic efficacy, or constituent-herb context.
Condition-aware causation represents adverse or causal relations constrained by disease or gene contexts, such as a drug causing a symptom only under a specific disease state or through a gene-related mechanism.
Condition-aware co-occurrence represents statistical associations observed within a defined population, such as symptom--disease or symptom--herb co-occurrence patterns derived from clinical records.

The contextual entity types in LingShu include Syndrome, Therapeutic Efficacy, Population, Gene, Disease, Chemical, and Herb, while herb-related contexts are further distinguished by contextual roles such as constituent herb, added herb, and removed herb.
Among them, Syndrome, Disease, Gene, Therapeutic Efficacy, and Chemical encode clinical or mechanistic constraints; Population encodes cohort-specific statistical contexts.
The herb-related contextual types are further distinguished according to prescription semantics.
Herb denotes an internal composition context, indicating that a prescription treats a symptom through one of its constituent herbs.
In contrast, HerbAdd and HerbRemove denote prescription-modification contexts.
HerbAdd indicates that a prescription can treat a symptom, syndrome, TCM Disease, or disease after an additional herb is added, whereas HerbRemove indicates that the therapeutic relation holds after a specific herb is removed.
This design enables LingShu to represent both general biomedical conditional knowledge and TCM-specific clinical rules such as syndrome differentiation, herb-mediated formula compatibility, and additive or subtractive prescription modification.

\subsection{Data Sources}

LingShu is constructed with the ISPO symptom ontology as its core, systematically integrating heterogeneous medical knowledge across multiple dimensions.
As summarized in Table \ref{tab:data_summary}, the data repository encompasses 4,490 Chinese clinical electronic medical records, 1,541 classical TCM texts, 435 modern TCM books, 6 medical ontologies or terminologies, and 11 biomedical and TCM databases.

\begin{table}[t]
  \centering
  \caption{Summary of multi-source heterogeneous data integrated in LingShu}
  \label{tab:data_summary}
  \begin{tabular}{p{0.2\textwidth}p{0.16\textwidth}p{0.55\textwidth}}
  \toprule
  \textbf{Data Category} & \textbf{Quantity} & \textbf{Description} \\
  \midrule
  Clinical Records & 4,490 Records & Inpatient records of COVID-19 (n=1,788) and Cirrhosis (n=2,702) from top-tier TCM hospitals. \\
  \midrule
  Classical TCM Texts & 1,541 Texts & Canonical works such as \textit{Ben Cao Chong Yuan}, \textit{Zhu Bing Yuan Hou Lun}, and \textit{Wen Bing Tiao Bian}. \\
  \midrule
  Modern TCM Books & 435 Books & \textit{Chinese Pharmacopoeia}, \textit{TCM Internal Medicine}, \textit{National Standard of Syndromes}, and other modern TCM references. \\
  \midrule
  Medical Ontologies & 6 Terminologies & ISPO, UMLS, ICD-11, MeSH, HPO, and SNOMED CT. \\
  \midrule
  Biomedical and TCM Databases & 11 Databases & TCMIO, HIT, SymMap, SympGAN, DrugBank, DisGeNET, MalaCards, STITCH, STRING, SIDER, and Orphanet. \\
  \midrule
  Biomedical Literature & 581,920 Abstracts & PubMed. \\
  \bottomrule
  \end{tabular}
\end{table}

\subsubsection{Classical TCM Texts}

Classical TCM texts were selected to cover canonical theory, materia medica, formula composition, syndrome differentiation, acupuncture, disease etiology, and historical clinical cases. We retained texts that contained extractable medical content relevant to the LingShu ontology and excluded duplicated editions, non-medical fragments, and records with insufficient textual quality after OCR and formatting checks. The final corpus contained 1,541 classical texts. Raw texts were cleaned to remove page headers, page numbers, editorial noise, OCR artifacts, garbled characters, and excessive line breaks, and were then segmented into evidence-preserving text blocks for ontology-constrained extraction.

At the entity level, classical TCM texts contributed the main TCM-specific concept classes in LingShu, including Symptom, Herb, Prescription, Syndrome, Therapeutic Efficacy, Therapeutic Method, TCM Disease, Meridian, and Acupoint.
These sources were therefore used not only as textual evidence for relation extraction, but also as the primary source for historical terminology variants, prescription composition, herb efficacies, syndrome manifestations, meridian localization, and acupoint-related therapeutic knowledge.

\subsubsection{Modern TCM Books}
Modern TCM books were selected from pharmacopoeias, national terminology standards, textbooks, clinical specialty references, formula books, and medical case collections.
These sources were included when they provided standardized or clinically interpretable descriptions of symptoms, syndromes, herbs, prescriptions, therapeutic methods, TCM diseases, meridians, or acupoints.
The final set contained 435 books, including the \textit{Chinese Pharmacopoeia}, \textit{National Standard of Syndromes}, and major TCM clinical textbooks. 
Compared with classical texts, modern sources were used preferentially for terminology standardization, symptom-cluster definitions, prescription composition, and contemporary clinical interpretation.

At the entity level, modern TCM books complemented classical texts by contributing Symptom, Symptom Cluster, Herb, Prescription, Syndrome, Therapeutic Efficacy, Therapeutic Method, TCM Disease, Meridian, Acupoint, and selected modern Disease concepts. 
They were especially important for standardizing TCM terminology, defining symptom clusters, aligning traditional disease names with contemporary clinical descriptions, and providing modern textbook-level evidence for prescription, syndrome, therapeutic-method, and specialty-specific clinical knowledge.

\subsubsection{Biomedical Ontologies and Terminologies}

Biomedical ontologies and terminologies were included to provide controlled vocabularies, synonyms, identifiers, and cross-resource alignment anchors for symptom-centered integration. We selected ISPO, UMLS, ICD-11, MeSH, HPO, and SNOMED CT because they cover complementary aspects of the LingShu ontology: ISPO provides TCM-oriented symptom phenotype concepts \cite{shu2024ispo}; UMLS and MeSH support biomedical terminology normalization \cite{bodenreider2004unified,lipscomb2000medical}; ICD-11 provides disease classification \cite{harrison2021icd}; HPO provides standardized phenotypic abnormality terms \cite{kohler2021human}; and SNOMED CT provides broad clinical terminology coverage \cite{donnelly2006snomed,schulz2023snomed}.

For each terminology source, we extracted preferred names, synonyms, source identifiers, cross-references, and hierarchy-like relations when available. Records were retained when they could be mapped to the predefined LingShu entity types, primarily Symptom and Disease. Terms outside the medical scope of LingShu, administrative concepts, overly broad categories, and records without usable names or identifiers were excluded. Each retained source record was stored as an atom and then mapped to a unified LingShu Concept ID through exact matching, synonym matching, cross-reference alignment, and manual or rule-based review for ambiguous cases. This process allowed LingShu to preserve source-level provenance while using biomedical terminologies as normalization anchors for symptom and disease concepts.

After normalization, the terminology sources contributed complementary concept anchors to LingShu. ISPO was used as the core TCM-oriented symptom phenotype reference. UMLS, MeSH, HPO, and SNOMED CT supported symptom and clinical phenotype normalization, while UMLS, MeSH, ICD-11, and SNOMED CT supported modern Disease concept alignment. These ontology-derived records were used mainly as concept-level normalization anchors rather than as direct sources of TCM disease names, therapeutic relations, or mechanistic associations.

\subsubsection{Biomedical and TCM Databases}

Structured biomedical and TCM databases were included to provide curated associations that are difficult to recover reliably from free text alone. We selected TCMIO~\cite{liu2020tcmio}, HIT 2.0~\cite{yan2022hit}, SymMap~\cite{wu2019symmap}, SympGAN~\cite{lu2023sympgan}, DrugBank~\cite{wishart2018drugbank}, DisGeNET~\cite{pinero2020disgenet}, MalaCards~\cite{rappaport2017malacards}, STITCH~\cite{kuhn2014stitch}, STRING~\cite{szklarczyk2019string}, SIDER~\cite{kuhn2016sider}, and Orphanet~\cite{weinreich2008orphanet} because together they cover herb--target, herb--symptom, herb--disease, drug--target, drug--drug, chemical--gene, gene--gene, disease--gene, symptom--disease, and side-effect associations. These sources were used as structured evidence for connecting TCM therapeutic knowledge with modern biomedical mechanisms.

Database records were processed using a source-specific parsing and normalization workflow. For each resource, we extracted entity names, source identifiers, synonyms, relation endpoints, relation labels, and evidence fields when available. Records were retained only when both relation endpoints could be mapped to LingShu entity types and when the relation semantics could be converted into one of the predefined LingShu triple or contextual quadruple schemas. Records with missing endpoints, unresolved entity types, unsupported relation semantics, or duplicated source-level entries were excluded or merged during normalization. 
At the entity and relation levels, 
SymMap contributed cross-paradigm records involving Symptom, Herb, Disease, Gene, and Chemical entities; 
TCMIO and HIT 2.0 contributed herb-related records, with HIT 2.0 further supporting Herb--Gene associations; 
DrugBank contributed Drug concepts and drug-related Disease, Symptom, and Gene associations; 
DisGeNET, MalaCards, and Orphanet contributed Disease, Symptom, and Gene records for disease genomics and symptom--disease integration; 
STITCH contributed Chemical and Gene records, 
STRING contributed Gene records and gene--gene associations, and 
SIDER contributed Chemical and Symptom records related to indications and side effects. 
These source-specific records were normalized into the LingShu entity space, converted into predefined triple or contextual quadruple schemas when their relation semantics were compatible, and linked to their source database names and identifiers for provenance tracking.

\subsubsection{Biomedical Literature}

Biomedical literature was included to supplement structured databases with sentence-level evidence for mechanism-oriented contextual relations. PubMed was used as the primary literature source because it provides broad coverage of biomedical abstracts and controlled indexing information \cite{canese2013pubmed}. Unlike terminology and database resources, PubMed records were not used as a static vocabulary source; instead, they were used to identify text-supported relations involving Disease, Symptom, Drug, and Gene entities.

Candidate abstracts and sentences were retrieved and filtered according to the LingShu biomedical entity vocabulary and the predefined contextual relation schemas. Candidate entity mentions were first identified by dictionary matching and distant supervision against normalized LingShu concepts. Sentences were retained when they contained at least two mappable biomedical entities and expressed a potential treatment, causation, adverse-effect, or mechanism-mediated relation. An LLM-assisted relation classification module was then applied to determine whether each candidate sentence supported a valid typed contextual quadruple, such as a drug treating a symptom under a disease context, a drug causing a symptom under a disease context, or a drug treating a disease through a gene-related mechanism. Relation candidates inconsistent with the LingShu schema or lacking sufficient sentence-level evidence were discarded. Each retained relation was stored with its source sentence and PubMed record information for evidence traceability.

In the source-to-entity mapping, PubMed contributed modern biomedical entity classes, including Disease, Symptom, Drug, and Gene. It was not used to expand TCM-specific entities such as Herb, Prescription, Syndrome, Meridian, or Acupoint. This separation ensured that PubMed-derived evidence was restricted to biomedical treatment, adverse-effect, and mechanism-oriented relations, while TCM-specific clinical and theoretical knowledge remained grounded in TCM texts, books, clinical records, and specialized TCM databases.

\subsubsection{Clinical Electronic Medical Records}

Clinical records were included to provide real-world evidence for symptom, syndrome, disease, herb, and population-context associations. The corpus contained 4,490 inpatient records, including 1,788 COVID-19 records from five top-tier TCM hospitals in Hubei Province and 2,702 cirrhosis records from Hubei Provincial Hospital of Traditional Chinese Medicine. Records were retained when they contained sufficient diagnostic, symptom, syndrome, prescription, and demographic information for patient-level extraction and cohort-level aggregation. Personal identifiers were removed before annotation. Patient-level entities were extracted with HCPSAS and manually reviewed, and cohort-level herb--symptom and symptom--disease co-occurrences were retained only after the predefined statistical testing procedure.

In terms of entity coverage, clinical EMRs contributed patient-level Symptom, Syndrome, Herb, and Population entities. Symptom, syndrome, and herb mentions were used for patient-level clinical annotation and normalization, whereas Population entities represented cohort-level contexts defined by the primary disease and demographic characteristics. Thus, clinical records primarily supported real-world symptom--syndrome--herb associations and population-context co-occurrence knowledge rather than broad biomedical vocabulary construction.

\begin{table*}[htbp]
  \centering
  \caption{Entity types and source in LingShu.}
  \label{tab:entity_source_mapping}
  \small
  \renewcommand{\arraystretch}{1.12}
  \begin{tabular}{p{0.2\textwidth}p{0.74\textwidth}}
  \toprule
  \textbf{Entity type} & \textbf{Sources} \\
  \midrule
  Symptom & Classical TCM texts, modern TCM books, SymMap, SympGAN, SIDER, DisGeNET, MalaCards, Orphanet, UMLS, MeSH, ISPO, clinical EMRs, HPO, package inserts \\
  Symptom Cluster & Modern TCM books \\
  Herb & Classical TCM texts, modern TCM books, SymMap, clinical EMRs, HIT 2.0, TCMIO, package inserts \\
  Prescription & Classical TCM texts, modern TCM books \\
  Chinese Patent Medicine & Modern TCM books, package inserts \\
  Syndrome & Classical TCM texts, modern TCM books, clinical EMRs, package inserts \\
  Therapeutic Efficacy & Classical TCM texts, modern TCM books, package inserts \\
  Therapeutic Method & Classical TCM texts, modern TCM books \\
  TCM Disease & Classical TCM texts, modern TCM books \\
  Meridian & Classical TCM texts, modern TCM books \\
  Acupoint & Classical TCM texts, modern TCM books \\
  Population & Clinical EMRs \\
  Disease & Modern TCM books, SymMap, SympGAN, DisGeNET, MalaCards, Orphanet, UMLS, MeSH, ICD-11, package inserts \\
  Gene & STITCH, SymMap, DisGeNET, STRING, MalaCards, Orphanet, DrugBank, HIT 2.0, SympGAN, PubMed \\
  Chemical & STITCH, SymMap, SIDER \\
  Drug & SympGAN, DrugBank, PubMed \\
  \bottomrule
  \end{tabular}
\end{table*}

Following resource-centric reporting practices in large-scale biomedical knowledge graph construction \cite{chandak2023building}, we treated each input as a traceable primary source rather than as an undifferentiated text or database collection. Data sources were selected according to four criteria: relevance to the LingShu ontology, authority or community use in the corresponding domain, availability of structured identifiers or recoverable textual evidence, and complementarity across TCM clinical knowledge, biomedical terminology, pharmacological mechanisms, and real-world clinical records. For each source, we recorded its source category, version or edition when available, retrieval or compilation procedure, selected fields, filtering rules, entity and relation types contributed to LingShu, and provenance information retained after integration.

\subsection{Knowledge Acquisition}

This subsection describes how structured knowledge was acquired from three major unstructured source categories: ancient and modern TCM books, clinical electronic medical records, and biomedical literature.
Because these sources differ substantially in linguistic style, structural regularity, and evidence type, we adopted source-specific extraction strategies under the unified LingShu ontology schema.
For TCM books, we developed an ontology-constrained LLM-based extraction agent to support book-scale extraction of TCM entities, triples, and contextual quadruples.
For clinical electronic medical records, we used the HCPSAS human--computer collaborative annotation system with embedded deep learning models and manual review to extract patient-level clinical entities.
For biomedical literature, we used LLM-assisted relation classification to identify mechanism-oriented contextual relations among diseases, symptoms, drugs, and genes.

\subsubsection{Atom--Concept Identifier Initialization}

Before large-scale extraction from unstructured text, LingShu established a type-specific concept identifier space through the integration and normalization of structured public resources, terminology dictionaries, and curated knowledge bases.
Source-specific entity records were retained as atoms, whereas semantically equivalent atoms were merged into unified concepts and assigned stable Concept IDs.
This atom--concept design preserves source-level provenance while supporting concept-level normalization, alignment, and reasoning.
The resulting Concept ID system served as the anchor vocabulary for subsequent entity alignment.
Entities extracted from unstructured text were first mapped to existing concepts whenever possible; only entities that could not be aligned to any existing concept were incrementally assigned new type-specific Concept IDs under the same identifier scheme.

\subsubsection{Ancient and Modern TCM Books}

For ancient and modern TCM books, we designed an ontology-constrained LLM-based extraction agent to acquire structured TCM knowledge from book-scale textual corpora.
The agent was organized around four functional components: text preprocessing, ontology-constrained joint extraction, tool-augmented entity anchoring and type verification, and closed-loop validation.
The extraction agent was implemented using Qwen3.5-27B~\cite{qwen3.5}, with prompts constrained by the predefined LingShu ontology and structured output requirements.
Because these sources primarily describe TCM theory, materia medica, prescriptions, syndrome differentiation, treatment principles, acupuncture-related knowledge, and clinical cases, the extraction scope was restricted to ten TCM entity types: Symptom, Herb, Prescription, Chinese Patent Medicine, Syndrome, Therapeutic Efficacy, Therapeutic Method, TCM Disease, Meridian, and Acupoint.
Modern biomedical entities, including Disease, Drug, Chemical, and Gene, were not extracted from this source category.

In the preprocessing component, raw book texts were first cleaned through rule-based procedures to remove non-medical noise, such as page headers, page numbers, redundant publication information, references, OCR artifacts, garbled characters, and excessive line breaks.
Long texts were then partitioned using sliding windows with overlap to preserve local context.
Based on these preliminary segments, the LLM further refined the texts into semantically self-contained knowledge blocks, such as prescription entries, herb descriptions, syndrome analyses, medical case fragments, and prescription modification instructions.
Necessary antecedent information, such as the prescription name, herb name, syndrome topic, or TCM disease name, was retained to ensure that each knowledge block could be independently used for relation extraction.

In the ontology-constrained extraction component, the agent jointly extracted entities, triples, and contextual quadruples from each knowledge block in a single reasoning step.
The extraction process was constrained by the predefined LingShu ontology, including entity type definitions, admissible triple patterns, and admissible contextual quadruple patterns.
For TCM books, the extraction module covered eight basic predicates: \textit{Treats}, \textit{Produces}, \textit{ConsistOf}, \textit{ManifestationOf}, \textit{BelongsTo}, \textit{Causes}, \textit{LocatedIn}, and \textit{AssociatedWith}.
These predicates were instantiated into 27 typed triple patterns among the ten TCM entity types.
For contextual knowledge, the agent directly extracted prescription-modification quadruples.
Specifically, prescriptions were linked to symptoms, syndromes, or TCM disease names under either herb-addition or herb-removal contexts, resulting in six typed contextual quadruple patterns.
These quadruples capture the clinical logic that a prescription may become applicable to a specific symptom, syndrome, or TCM disease after adding or removing a particular herb.

In the tool-augmented entity anchoring and type verification component, the agent invoked external tools to reduce ambiguity in entity normalization and type assignment.
For entity anchoring, candidate entities were aligned to the Concept ID system described above through a three-stage matching procedure: exact name matching against existing graph vocabularies, normalized name matching after removing irrelevant modifiers and surface-form variation, and vector-based semantic retrieval for unresolved entities.
For each candidate, the resolver returned possible matched concepts and their synonyms, allowing the agent to determine whether the entity should be mapped to an existing concept.
Entities that could not be aligned to existing concepts were incrementally assigned new type-specific Concept IDs under the established LingShu identifier scheme, rather than being assigned arbitrary identifiers by the LLM.

For entity type verification, the agent could query the existing concept coding table constructed from public structured resources.
When the LLM was uncertain whether a candidate term belonged to a specific entity type, the tool returned reference concepts, their assigned entity types, and synonym evidence from the existing Concept ID system.
This allowed the agent to make type decisions with support from the established ontology and concept inventory, rather than relying solely on the generative model.
The verified entity types were then used as constraints for subsequent triple and quadruple validation.

In the closed-loop validation component, a deterministic validator was applied to each agent output to ensure structural and ontological consistency.
The validator checked entity type validity, identifier consistency, relation endpoint existence, head--tail type compatibility for triples, and head--tail--context type compatibility for contextual quadruples.
When validation failed, the error messages were returned to the agent as structured feedback, and the agent regenerated the corrected output.
This generation--validation--feedback loop was repeated for up to three rounds.
After validation, missing relation endpoints explicitly present in extracted relations but absent from the entity list were identified and submitted to a targeted entity-completion module.
The completed entities were aligned, merged back into the entity list, and revalidated.
The final output of this module consisted of validated entities, triples, and prescription-modification quadruples linked to their source books and knowledge blocks, providing traceable structured knowledge for subsequent post-verification and graph construction.

\subsubsection{Clinical Electronic Medical Records}

For clinical electronic medical records, we used HCPSAS \cite{zou2021phenonizer}, a human--computer collaborative annotation system developed by our group, to extract patient-level clinical entities from real-world TCM records.
HCPSAS embeds deep learning models to assist manual annotation and review, supporting the identification of fine-grained symptom entities, including positive symptoms, negative symptoms, and symptom trajectories.
Human annotators then corrected and reviewed the model-assisted annotations to ensure the reliability of clinical entity extraction.
Herb, disease, and syndrome entities were further extracted and standardized through a combination of dictionary matching, rule-based recognition, and manual verification.

To construct population-context co-occurrence knowledge, we defined each population entity as a disease-specific patient cohort characterized by the primary disease and demographic attributes such as age distribution and sex ratio.
For example, a population entity may correspond to a cirrhosis or COVID-19 cohort with cohort-level age and sex statistics.
Within each cohort, patient-level herb--symptom co-occurrences were aggregated. Associations passing the co-occurrence significance testing procedure described by Gan et al. \cite{gan2023network} were retained as population-context co-occurrence knowledge.
The retained associations were represented as contextual quadruples in the form $(h, r, t, c)$, where the symptom and herb constitute the co-occurrence relation and the population entity serves as the context that constrains the validity of the association.

\subsubsection{Biomedical Literature}

For biomedical literature, we focused on extracting mechanism-oriented contextual relations involving Disease, Symptom, Drug, and Gene entities.
Randomized controlled trials and related biomedical abstracts were retrieved from PubMed as evidence sources.
Candidate entity mentions and relation instances were first identified through distant supervision and dictionary-based matching against the biomedical vocabularies integrated in LingShu.
An LLM-assisted relation classification module was then used to determine whether each candidate sentence expressed a valid contextual relation under the predefined LingShu schema.

The classification task focused on distinguishing treatment, causation, and mechanism-mediated semantics.
For example, the model determined whether a drug treated a symptom under a disease context, caused a symptom as an adverse effect under a disease context, treated a symptom through a gene-related mechanism, caused a symptom through a gene-related mechanism, or treated a disease through a gene-related mechanism.
Only relation instances consistent with the typed contextual quadruple schema were retained.
Each retained relation was linked to its source sentence to ensure evidence traceability, and the extracted candidates were further reviewed before being incorporated into the graph construction pipeline.

\subsection{Terminology Normalization and Entity Alignment}

Building on the Atom--Concept identifier system, terminology normalization and entity alignment were performed to represent heterogeneous source terms and extracted mentions under a unified concept-level schema.
This process allowed LingShu to preserve source-level provenance while supporting concept-level retrieval, reasoning, and cross-source integration.

For TCM entities, normalization was conducted according to type-specific lexical and semantic rules.
Syndrome expressions were normalized by considering their compositional structure, which typically follows the pattern of etiology, pathological location, and pathological state.
Synonymous expressions, historical variants, and alternative translations were mapped to unified syndrome concepts through exact matching, rule-based synonym expansion, and concept validation based on standard TCM terminology resources.
Herb entities were normalized using controlled vocabularies derived from authoritative resources such as the Chinese Pharmacopoeia and TCM databases, with aliases, processed forms, and regional variants aligned to unified herb concepts.
Prescription entities were normalized by combining name-based synonym matching with composition-based verification, so that formula names were merged only when their core herbal compositions were consistent.
Therapeutic Efficacy and Therapeutic Method terms were further standardized by decomposing compound functional expressions into more atomic therapeutic units when appropriate.
Other TCM entity types, including symptoms, TCM disease names, meridians, and acupoints, were normalized using domain dictionaries, ontology constraints, and concept-level consistency checks.

For modern biomedical entities, alignment relied on cross-source identifiers, curated mappings, and synonym relationships from integrated ontologies and databases.
Disease concepts were aligned across resources such as UMLS, MeSH, MalaCards, DisGeNET, ICD-11, Orphanet, and SymMap.
Because cross-references among these resources are sometimes incomplete or inconsistent, mappings were filtered to retain high-confidence alignments, and finer-grained concepts were preserved when different resources represented the same surface term at different levels of specificity.
Cross-lingual synonym relationships from Chinese and English resources were also used to expand alignments without changing the underlying concept identifiers.
Analogous procedures were applied to chemicals, drugs, and genes, with gene entities consistently linked to standard gene identifiers.

Entities extracted from unstructured text were subsequently anchored to this normalized concept inventory.
The alignment process combined exact name matching, normalized name matching, and vector-based semantic retrieval.
Exact matching was used when an extracted mention directly matched an existing concept name or synonym.
Normalized matching reduced surface-form variation by removing irrelevant modifiers, punctuation, and formatting differences.
For unresolved mentions, vector-based retrieval was used to identify semantically similar candidate concepts from the existing concept inventory.
When an extracted entity could not be aligned to any existing concept, it was assigned a new type-specific Concept ID under the established LingShu identifier scheme.
All mappings were constrained by the ontology-defined entity types, ensuring that concept assignment and relation construction remained consistent with the schema.

\subsection{Post-verification and Quality Control}

To ensure that large-scale extracted knowledge could be reliably incorporated into LingShu, we introduced an ontology-guided post-verification procedure after entity normalization and alignment. This procedure aimed to improve three aspects of data quality: the semantic validity of extracted entities, the consistency of entity type assignments, and the schema compatibility of triples and contextual quadruples. Rather than relying directly on raw extraction outputs, post-verification used the LingShu ontology, the normalized Concept ID system, and source-context evidence to refine extracted knowledge before graph construction.

The first stage focused on source-grounded preparation. Extracted entities and relations were linked back to their original textual evidence, allowing later verification to be conducted with contextual support rather than isolated entity names. Concept assignments were also examined for potential inconsistencies, especially cases in which the same surface expression was associated with multiple concepts or entity types.

The second stage performed entity-level semantic verification. Candidate entities collected from both entity outputs and relation endpoints were evaluated against the LingShu ontology and the normalized concept inventory. Candidates that matched existing concepts were mapped to their corresponding Concept IDs. Ambiguous or previously unseen candidates were further checked using their source context to determine whether they represented valid medical entities, required type correction, should be decomposed into multiple entities, or should be excluded as non-medical terms, vague references, or over-extended text fragments.

The third stage performed relation-level schema verification. Each triple was retained only when both endpoints could be mapped to valid concepts and their entity types satisfied the corresponding typed relation pattern. For contextual quadruples, the same verification was applied to the head, tail, and contextual entities, and a quadruple was retained only when all participating entity types were compatible with the predefined contextual schema. When an entity mention could correspond to multiple valid types, relation instances were expanded only over schema-compatible type combinations.

Through this post-verification procedure, LingShu reduced noise introduced during large-scale extraction and ensured that graph construction was based on semantically valid entities and ontology-consistent relations.

\subsection{Logical Inference over Verified Relations}

After extraction, normalization, entity alignment, and post-verification were completed, we performed rule-based logical inference over the verified relation set.
This step was not part of raw text extraction; instead, it operated on concept-level triples and contextual quadruples that had already passed ontology-based validation.
The purpose was to improve the contextual completeness and graph connectivity of LingShu by deriving additional schema-consistent relations from existing verified knowledge.

The inference rules were defined according to the LingShu ontology and executed deterministically.
First, verified triples were composed into contextual quadruples when they shared compatible entities and satisfied predefined semantic patterns.
For example, if $(h,\textit{Treats},s)$ and $(c,\textit{ManifestationOf},s)$ were both verified and their entity types matched an admissible contextual schema, LingShu derived $(h,\textit{Treats},s,c)$ as a syndrome-contextualized therapeutic quadruple. 
Similar rules were used to derive therapeutic-efficacy-, chemical-, gene-, and herb-composition-contextualized therapeutic knowledge when the corresponding supporting relations were present.
This enabled LingShu to derive context-aware therapeutic knowledge, such as herb--symptom treatment under a syndrome context or prescription--symptom treatment under a therapeutic efficacy or herb-composition context.

Second, selected contextual quadruples were projected into triples when the contextual relation entailed a valid binary relation under the ontology.
For instance, a contextual treatment assertion involving a prescription, symptom, and syndrome could support a corresponding prescription--symptom therapeutic triple, while a gene- or disease-context drug relation could support a binary drug--disease or drug--symptom relation when permitted by the typed triple schema.
This projection increased graph connectivity and supported downstream retrieval and reasoning over both binary and contextual relation forms.

All inferred relations were rechecked against the predefined typed triple and contextual quadruple schemas.
Only relations whose participating entity types satisfied the corresponding ontology constraints were retained.
Through this post-processing inference step, LingShu connected directly extracted knowledge, statistically derived clinical associations, and structured database knowledge into a unified relation system containing both triples and contextual quadruples.

\subsection{Knowledge Graph Storage}

After extraction, normalization, post-verification, and logical inference, the verified knowledge was stored in Neo4j using the Property Graph model.
Graph construction was performed primarily at the concept level: each normalized concept was represented as a node with its Concept ID, entity type, preferred name, synonyms, and source-level atom mappings.
This design preserved provenance from heterogeneous sources while supporting concept-level retrieval and reasoning.

Typed triples were stored as directed edges between head and tail concept nodes.
Each edge was labeled by its semantic predicate and constrained by the corresponding typed relation pattern $(h_{type}, r, t_{type})$ defined in the LingShu ontology.
For contextual quadruples $q=(h,r,t,c)$, the head and tail concepts were also connected by a directed edge, while the contextual entity $c$ was stored as structured relationship attributes, including its Concept ID, entity type, and contextual role.
This property-based representation enables LingShu to encode n-ary contextual knowledge without changing the underlying graph topology.

To support evidence traceability, each stored relation retained source information, including the data source, source identifier, and text evidence when available.
When multiple sources supported the same concept-level relation, their evidence records were preserved while the relation itself was represented in a unified schema.
The resulting graph database supports Cypher-based retrieval, subgraph exploration, contextualized relation querying, and downstream reasoning over both triples and contextual quadruples.

\section{Results}

We report the construction of LingShu at three levels: book-scale knowledge acquisition and post-verification, graph-wide coverage of entities and relations, and representative graph-enabled use cases.

\subsection{Book-scale Extraction and Post-verification Results}

The ontology-constrained extraction agent converted ancient and modern TCM books into structured candidate entities and relations.
Extraction covered ten TCM entity types, including symptoms, herbs, prescriptions, Chinese patent medicines, syndromes, therapeutic efficacies, therapeutic methods, TCM disease names, meridians, and acupoints.
At the relation level, the extracted instances were organized under 27 typed triple patterns derived from eight basic predicates, together with six typed contextual quadruple patterns for prescription modification under herb-addition or herb-removal contexts.

After initial extraction and aggregation, 1,167,163 unique candidate entities were submitted to entity-level verification.
Among them, 253,007 entities were excluded, corresponding to a removal rate of 21.7\%.
The excluded entities mainly included non-medical expressions, vague references, overly long text fragments, system or theory names, incorrectly typed entities, and compound expressions requiring decomposition.
After verification, 914,156 unique entities were retained and normalized to concept-level mappings.
Considering normalized-name expansion, compound-term decomposition, and multi-source provenance, the entity verification stage produced approximately 1.12 million concept mappings and 16.06 million source-grounded entity records.

Relation-level verification further removed invalid or dangling relations caused by unverified entity endpoints or schema-incompatible type combinations.
For triples, 12,179,458 extracted records were examined, of which 1,339,699 were removed.
The retained 10,839,759 triples were used as verified book-derived binary relations.
For contextual quadruples, 773,810 extracted records were examined, of which 151,470 were removed.
The retained 622,340 contextual quadruples mainly captured prescription-modification knowledge involving herb-addition and herb-removal contexts.
The entity- and relation-level post-verification results are summarized in Table \ref{tab:book_post_verification}.

\begin{table*}[htbp]
  \centering
  \caption{Post-verification results for book-scale knowledge extraction.}
  \label{tab:book_post_verification}
  \begin{tabular}{ccccc}
  \toprule
  \textbf{Verification Stage} & \textbf{Input} & \textbf{Removed} & \textbf{Retained} & \textbf{Removal Rate} \\
  \midrule
  Entity & 1,167,163 & 253,007 & 914,156 & 21.7\% \\
  \midrule
  Triple & 12,179,458 & 1,339,699 & 10,839,759 & 11.0\% \\
  \midrule
  Quadruple & 773,810 & 151,470 & 622,340 & 19.6\% \\
  \bottomrule
  \end{tabular}
\end{table*}

These results indicate that entity- and relation-level post-verification substantially reduced noise introduced during large-scale book extraction.
Entity-level verification removed more than one fifth of unique candidate entities before graph construction.
Relation-level verification further eliminated dangling or schema-incompatible triples and contextual quadruples caused by invalid entity endpoints or incompatible type combinations.
The verified book-derived entities and relations formed a major TCM-domain component of LingShu.

\subsection{Overview of Entities, Triples, and Quadruples}

After integrating structured biomedical resources, clinical electronic medical records, biomedical literature, and verified book-derived knowledge, LingShu was constructed as a large-scale contextualized knowledge graph spanning both TCM and modern biomedicine.
The final graph covers 16 entity types, including symptoms, symptom clusters, herbs, prescriptions, Chinese patent medicines, syndromes, therapeutic efficacies, therapeutic methods, TCM disease names, modern diseases, drugs, chemicals, genes, meridians, acupoints, and population entities.
Among them, three Population entities were represented as cohort-level contextual entities for population-specific quadruples, whereas Table \ref{tab:lingshu_statistics} reports the 15 non-Population biomedical and TCM entity categories.
At the schema level, LingShu defines 64 typed triple relation patterns and 35 typed contextual quadruple relation patterns.
Instances under these relation patterns were generated through direct extraction and ontology-guided logical inference over verified triples and contextual quadruples.

Across the 16 entity types, LingShu contains approximately 17.33 million atom-level entity records, which are normalized into 1,035,937 unified concepts.
This atom--concept design allows the graph to preserve source-level provenance while supporting concept-level retrieval, integration, and reasoning.
Detailed statistics for each entity category are provided in Table \ref{tab:lingshu_statistics}.

All statistics reported in this section correspond to a fixed exported version of LingShu used for manuscript analysis, although the online platform is continuously updated through ongoing human-in-the-loop curation.

\begin{table}[htbp]
  \centering
  \caption{Statistics of the 15 core entity categories in LingShu. Population entities were used as contextual entities for population-specific quadruples and were not included in this table. "Atom Count" refers to the raw count of entities (AtomId) collected from multiple sources; "Concept Count" refers to the unique normalized nodes (ConceptId) in the final graph.}
  \label{tab:lingshu_statistics}
  \begin{tabular}{lcc}
  \toprule
  \textbf{Entity Type} & \textbf{Atom Count} & \textbf{Concept Count} \\
  \midrule
  Herb                & 5,689,876 & 34,503 \\
  Symptom             & 5,047,386 & 215,145 \\
  Therapeutic Efficacy & 1,744,571 & 49,725 \\
  Syndrome            & 1,230,921 & 77,693 \\
  Prescription        & 1,117,106 & 126,361 \\
  TCM Disease          & 1,121,078 & 31,260 \\
  Disease             & 600,609   & 172,392 \\
  Chemical            & 386,359   & 228,672 \\
  Gene                & 154,128   & 62,763 \\
  Acupoint            & 70,492    & 1,556 \\
  Therapeutic Method  & 82,012    & 5,500 \\
  Symptom Cluster     & 6,387     & 6,387 \\
  Drug                & 43,111    & 14,615 \\
  Meridian            & 24,433    & 307 \\
  Chinese Patent Medicine & 11,752 & 9,058 \\
  \midrule
  \textbf{Total} & \textbf{17,330,221} & \textbf{1,035,937} \\
  \bottomrule
  \end{tabular}
\end{table}

In terms of binary semantic connectivity, LingShu contains 17,185,769 verified triples distributed across the 64 typed triple relation patterns.
These triples cover therapeutic, causal, compositional, manifestation, interaction, association, hierarchy, indication, side-effect, localization, and therapeutic-method membership semantics.
For example, they encode relations such as prescriptions composed of herbs, herbs producing therapeutic efficacies, syndromes manifested by symptoms, drugs interacting with other drugs, and symptoms associated with genes or diseases.
The distribution of typed triple relations is reported in Table \ref{tab:lingshu_triples}.

\begin{table}[htbp]
  \centering
  \caption{Statistics of atom-level semantic triples in LingShu with counts greater than 1,000.}
  \label{tab:lingshu_triples}
  \begin{tabular}{lllr}
      \toprule
      \textbf{Head Entity} & \textbf{Relation} & \textbf{Tail Entity} & \textbf{Count} \\ 
      \midrule
      Prescription & consists of & Herb & 3,360,533 \\ 
      Drug         & interacts with & Drug & 2,767,544 \\ 
      Prescription & treats & Symptom & 1,515,599 \\ 
      Herb         & produces & Therapeutic Efficacy & 1,442,841 \\ 
      Herb         & treats & Symptom & 923,570 \\ 
      Gene         & interacts with & Gene & 840,513 \\ 
      Syndrome     & manifests as & Symptom & 815,784 \\ 
      Chemical     & interacts with & Chemical & 516,834 \\ 
      Prescription & produces & Therapeutic Efficacy & 509,846 \\ 
      Chemical     & associated with & Gene & 500,327 \\ 
      Symptom      & associated with & Gene & 414,248 \\ 
      Prescription & treats & Syndrome & 392,011 \\ 
      Symptom      & causes & TCM Disease & 385,414 \\ 
      Prescription & treats & TCM Disease & 338,646 \\ 
      Herb         & treats & TCM Disease & 335,991 \\ 
      Disease      & associated with & Gene & 317,227 \\ 
      Symptom      & associated with & Disease & 288,360 \\ 
      Herb         & treats & Syndrome & 252,517 \\ 
      Disease      & is a & Disease & 196,109 \\ 
      Herb         & associated with & Gene & 146,497 \\ 
      Symptom      & has side effect & Chemical & 128,846 \\ 
      Chinese Patent Medicine & consists of & Herb & 84,344 \\ 
      Symptom Cluster & consists of & Symptom & 76,041 \\ 
      Herb         & contains & Chemical & 69,766 \\ 
      Drug         & causes & Symptom & 65,745 \\ 
      Acupoint     & treats & Symptom & 57,790 \\ 
      Therapeutic Efficacy & treats & Symptom & 57,467 \\ 
      Therapeutic Efficacy & treats & Syndrome & 44,712 \\ 
      Chinese Patent Medicine & treats & Symptom & 39,161 \\ 
      Syndrome     & causes & TCM Disease & 33,758 \\ 
      Symptom      & associated with & Symptom & 23,912 \\ 
      Chinese Patent Medicine & produces & Therapeutic Efficacy & 23,468 \\ 
      Drug         & associated with & Gene & 22,038 \\ 
      Drug         & treats & Disease & 20,839 \\ 
      Chinese Patent Medicine & treats & Disease & 20,544 \\ 
      Therapeutic Efficacy & treats & TCM Disease & 19,544 \\ 
      Chinese Patent Medicine & treats & Syndrome & 13,930 \\ 
      Drug         & treats & Symptom & 12,761 \\ 
      Therapeutic Method & treats & Symptom & 12,063 \\ 
      Acupoint     & treats & TCM Disease & 11,310 \\ 
      Acupoint     & produces & Therapeutic Efficacy & 10,523 \\ 
      Prescription & treats & Disease & 8,850 \\ 
      Prescription & belongs to & Therapeutic Method & 8,086 \\ 
      Syndrome     & located in & Meridian & 7,625 \\ 
      Symptom      & is a & Symptom & 6,852 \\ 
      Syndrome     & manifests as & Symptom Cluster & 6,591 \\ 
      Prescription & consists of & Prescription & 5,268 \\ 
      Therapeutic Method & treats & TCM Disease & 4,558 \\ 
      Therapeutic Method & treats & Syndrome & 4,400 \\ 
      Symptom      & causes & Disease & 4,313 \\ 
      Herb         & treats & Disease & 3,656 \\ 
      Drug         & causes & Disease & 3,378 \\ 
      Prescription & consists of & Acupoint & 3,357 \\ 
      Symptom      & has indication & Chemical & 2,750 \\ 
      Acupoint     & treats & Syndrome & 2,636 \\ 
      Chinese Patent Medicine & treats & TCM Disease & 2,344 \\ 
      \bottomrule
  \end{tabular}
\end{table}

Beyond triples, LingShu contains 22,286,936 contextualized quadruples distributed across 35 typed contextual quadruple relation patterns.
These quadruples were obtained through multiple routes, including direct extraction from TCM books, statistically supported population-context associations from clinical records, LLM-assisted extraction from biomedical literature, and ontology-guided logical inference over verified relations.
They explicitly encode contextual variables such as syndromes, therapeutic efficacies, molecular targets, chemical components, herb composition, herb addition or removal, disease contexts, and population cohorts.
The distribution of typed contextual quadruples is shown in Table \ref{tab:lingshu_quadruples}.
Compared with conventional triples, these contextualized quadruples specify the contextual entity associated with each relation, indicating the condition under which the association was recorded or inferred.

\begin{table}[htbp]
  \centering
  \caption{Statistics of atom-level contextualized quadruples in LingShu with counts greater than 1,000.}
  \label{tab:lingshu_quadruples}
  \begin{tabular}{llllr}
      \toprule
      \textbf{Head Entity} & \textbf{Relation} & \textbf{Tail Entity} & \textbf{Condition} & \textbf{Count} \\ 
      \midrule
      Prescription & treats & Symptom & Constituent herb & 12,952,727 \\
      Herb & treats & Symptom & Syndrome & 4,413,376 \\
      Prescription & treats & Symptom & Syndrome & 4,172,966 \\
      Prescription & treats & Symptom & Added herb & 356,930 \\
      Prescription & treats & Syndrome & Added herb & 137,566 \\
      Herb & treats & Symptom & Gene & 62,102 \\
      Prescription & treats & TCM Disease & Added herb & 51,969 \\
      Prescription & treats & Symptom & Removed herb & 40,551 \\
      Symptom & co-occurs with & Herb & Population & 36,237 \\
      Prescription & treats & Syndrome & Removed herb & 13,777 \\
      Drug & treats & Disease & Gene & 11,170 \\
      Symptom & co-occurs with & Disease & Population & 9,371 \\
      Prescription & treats & TCM Disease & Removed herb & 7,759 \\
      Drug & treats & Symptom & Disease & 6,758 \\
      Drug & causes & Symptom & Disease & 4,948 \\
      Drug & treats & Symptom & Gene & 3,423 \\
      Prescription & treats & Disease & Added herb & 1,484 \\
      Drug & causes & Symptom & Gene & 1,388 \\
      \bottomrule
  \end{tabular}
\end{table}

\subsection{Applications and Use Cases}

To support practical exploration and empirical research, we developed the LingShu web platform, accessible at \url{http://www.tcmkg.com/}.
The platform was developed to demonstrate how LingShu can be queried, inspected, and used for evidence-grounded exploration of symptom-centred, context-dependent medical knowledge.
These functions enable users to explore cross-paradigm associations among symptoms, syndromes, herbs, prescriptions, TCM disease names, modern diseases, drugs, genes, chemicals, acupoints, meridians, and molecular mechanisms.
The platform is designed for research, knowledge exploration, and decision-support purposes; its outputs do not constitute medical diagnosis, prescription, or treatment advice.

\subsubsection{Platform Overview}

The LingShu platform is organized around graph storage, knowledge retrieval, reasoning services, question answering, and user interaction.
At the storage layer, Neo4j manages concept-level entities, typed triples, and contextual quadruples, supporting multi-hop traversal and context-aware relation queries.
At the service layer, graph query, retrieval-augmented generation (RAG), and graph-based reasoning modules are encapsulated as callable interfaces.
The reasoning services can be used directly through the knowledge reasoning page and can also be invoked as tool functions by the multi-mode Q\&A agent.
At the interaction layer, the platform provides interfaces for data browsing, graph visualization, evidence inspection, multi-mode question answering, reasoning task configuration, and knowledge curation.

\subsubsection{Graph-based Knowledge Reasoning}

The knowledge reasoning module performs structured association inference over the constructed triple and quadruple networks using the unified concept layer.
It is built on Neo4j and the normalized \textit{concept\_id} system, so that reasoning is conducted over concept-level entities rather than heterogeneous source-specific names.
The module is organized into four exploratory scenarios: symptom-driven association analysis, disease-driven association analysis, syndrome--prescription--herb mechanism exploration, and acupuncture-related candidate prioritization.

In the symptom-driven scenario, the system recommends ranked candidate diseases, syndromes, herbs, and prescriptions from a user-provided symptom set.
In the disease-driven scenario, the system starts from a modern disease concept and retrieves related symptoms, syndromes, herbs, and prescriptions.
In the syndrome--prescription--herb mechanism scenario, the system prioritizes candidate gene targets associated with herbs, syndromes, or symptoms, supporting mechanism-oriented analysis of TCM therapeutic knowledge.
In the acupuncture-related scenario, the system supports symptom-to-acupoint candidate prioritization and gene--target exploration from acupoints or meridians.
Together, these functions constitute 14 structured reasoning modes across different head--tail entity type combinations.

The reasoning engine is implemented through a unified generic recommendation framework with two complementary strategies: PageRank-based ranking and path-based reasoning.
The PageRank-based strategy uses personalized PageRank over a unified graph projection to rapidly identify highly associated candidate entities, making it suitable for large-scale interactive candidate prioritization.
The path-based strategy searches for semantic paths connecting source and target entities and scores candidates by considering path length, source coverage, and graph connectivity.
For multi-symptom or multi-disease inputs, the path-based method can require a candidate target to be connected with all selected source concepts.
It also returns complete entity--relation alternating reasoning chains, enabling users to inspect the semantic basis of each result.

\begin{figure}[htbp]
  \centering
  \includegraphics[width=0.8\textwidth]{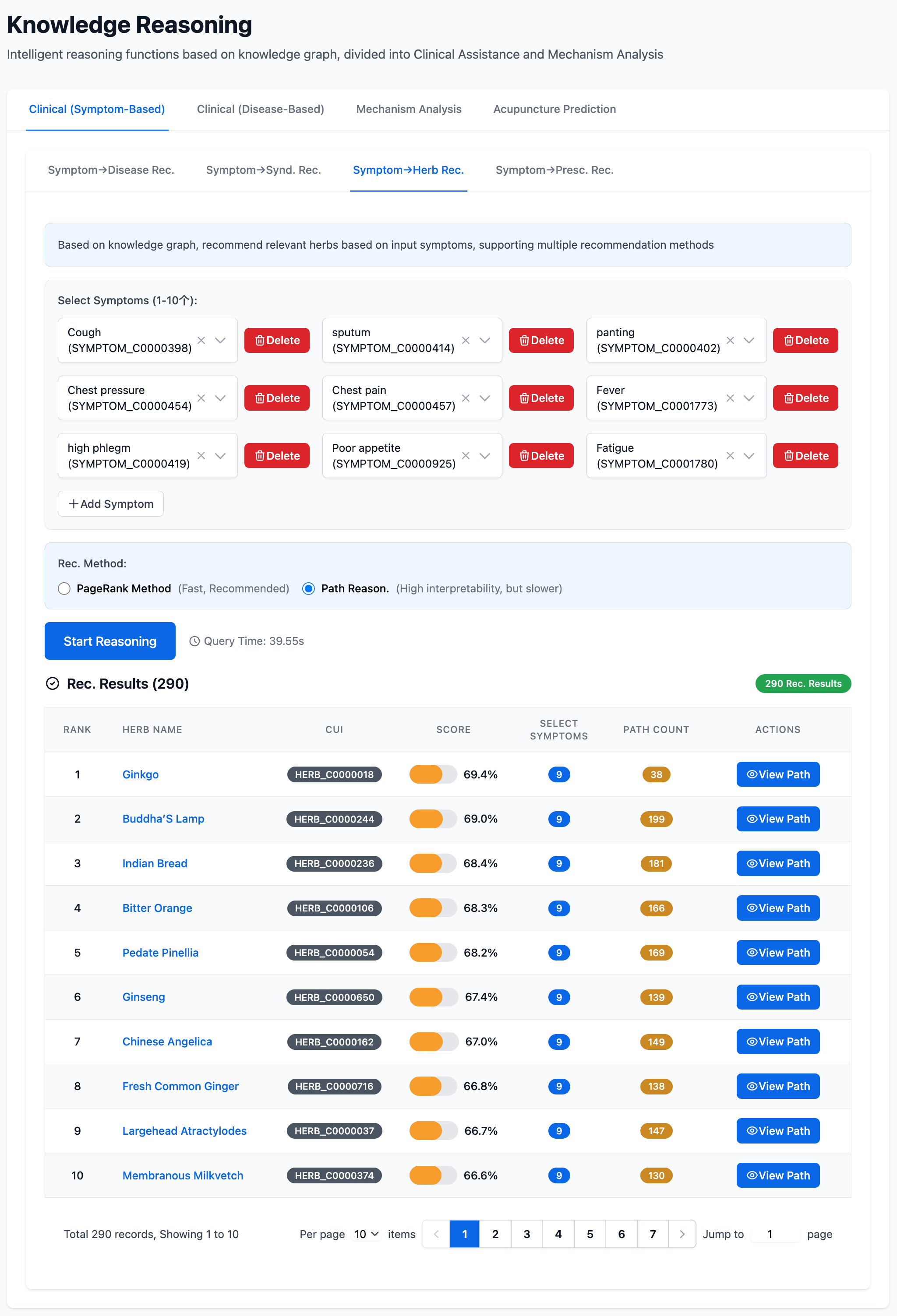}
  \caption{Example of symptom-driven herb recommendation using the graph-based reasoning engine.}
  \label{fig:kg_reasoning}
\end{figure}

\begin{figure}[htbp]
  \centering
  \includegraphics[width=0.8\textwidth]{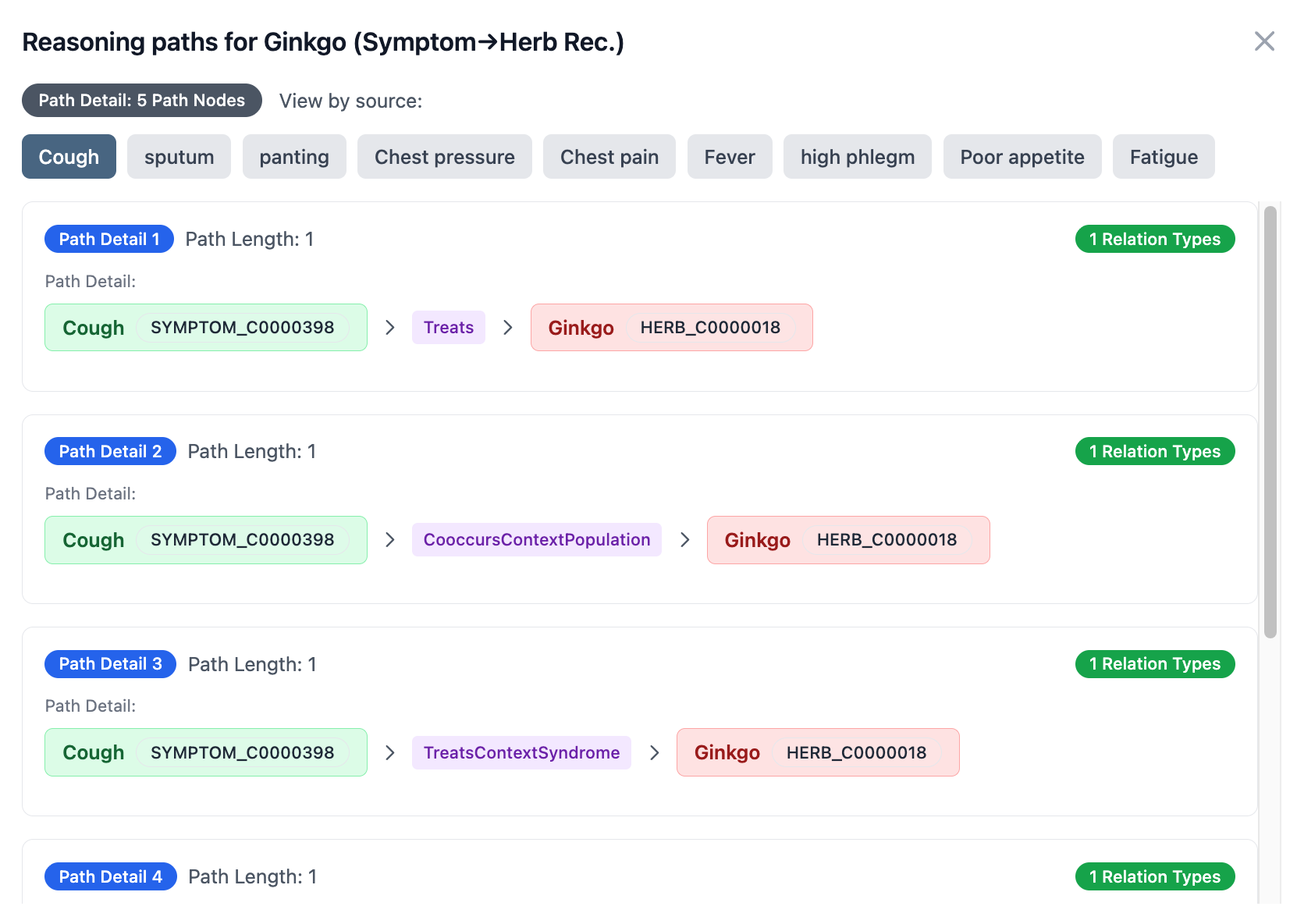}
  \caption{Visualized semantic reasoning paths connecting input symptoms to a recommended herb.}
  \label{fig:kg_reasoning_detail}
\end{figure}

As illustrated in Fig. \ref{fig:kg_reasoning}, the symptom-driven reasoning interface allows users to input multiple symptom concepts and obtain ranked candidate entities, such as herbs, using either PageRank-based ranking or path-based reasoning.
For each recommended result, the system reports ranking scores, path counts, and query information to support result inspection.
Fig. \ref{fig:kg_reasoning_detail} further shows the visualized semantic reasoning paths behind a recommended herb.
These paths expose the intermediate entities and relations connecting the input symptoms to the candidate result, thereby linking algorithmic scores with interpretable graph evidence.

The front-end reasoning interface supports entity search and multi-selection of up to ten source concepts, method switching between PageRank and path reasoning, paginated result display, ranking score, path count, query time, reasoning path visualization, and one-click navigation to the graph browsing interface.
By connecting graph structure, algorithmic scores, and explicit semantic paths, the reasoning module provides a verifiable entry point for syndrome differentiation, prescription and herb recommendation, disease--syndrome association analysis, and mechanism exploration involving herbs, syndromes, acupoints, meridians, and gene targets.

\begin{figure}[htbp]
  \centering
  \includegraphics[width=0.8\textwidth]{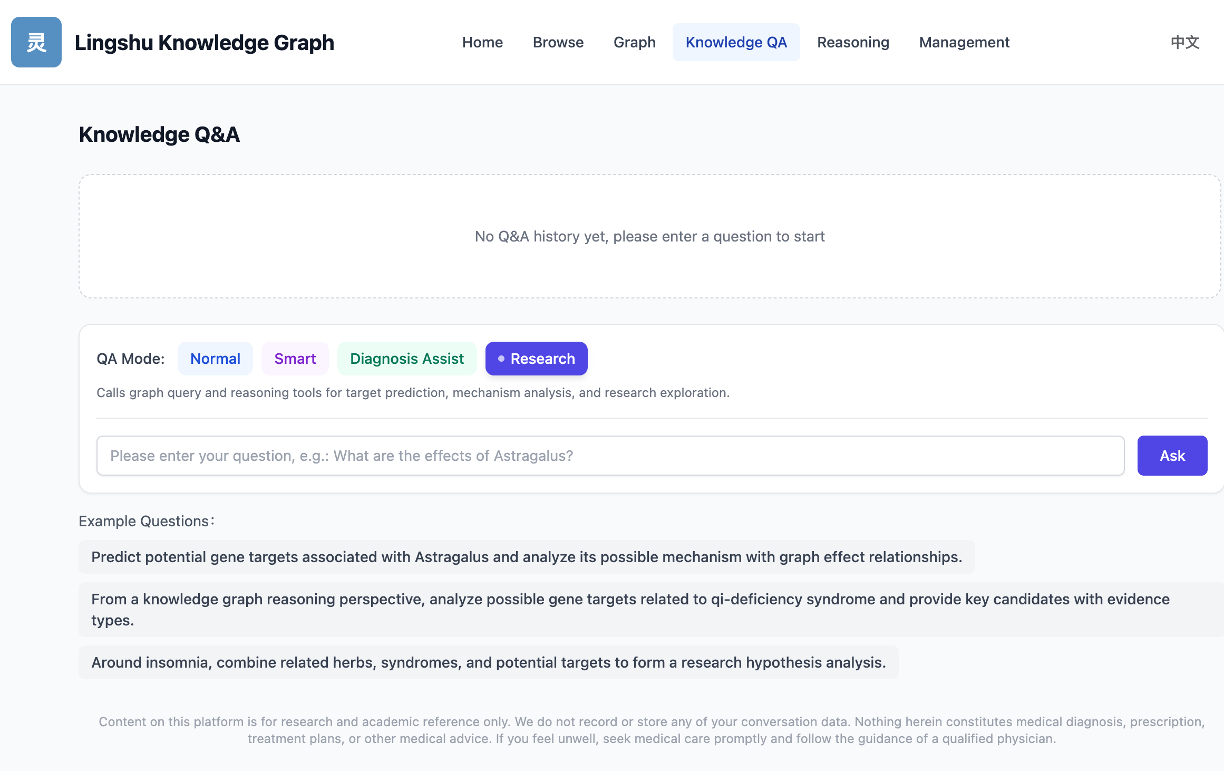}
  \caption{Multi-mode knowledge Q\&A interface supported by graph retrieval and reasoning tools.}
  \label{fig:kg_qa}
\end{figure}

\subsubsection{Multi-mode Knowledge Q\&A}

Built on the graph retrieval and reasoning services described above, the Knowledge Q\&A module provides a natural-language interface to LingShu.
As shown in Fig. \ref{fig:kg_qa}, the current system supports four interaction modes: Normal Mode, Smart Mode, Clinical Assistance Mode, and Research Mode.
These modes differ in the depth of reasoning, the tools invoked, and the intended use scenario.

Normal Mode is designed for direct knowledge retrieval and evidence-grounded answering.
Given a natural-language question, the system uses an LLM to generate structured SQL or Cypher queries, retrieves relevant knowledge from the graph database, and streams a Markdown-formatted answer to the user.
The structured query evidence is displayed together with the generated response, allowing users to inspect the basis of the answer.

Smart Mode is designed for complex questions requiring multi-step reasoning.
It follows a ReAct-style reasoning process \cite{yao2022react}, in which the agent alternates among reasoning, action, and observation steps.
During this process, the agent can call triple and contextual quadruple query tools, identify relevant concepts, retrieve schema-compatible relations, and synthesize an evidence-grounded answer.
The complete reasoning chain can be displayed in a collapsible format, making the intermediate reasoning process transparent for complex relation analysis.

Clinical Assistance Mode supports decision-support exploration based on free-text clinical descriptions.
Users can input clinical text such as chief complaint, present illness, symptoms, tongue description, and pulse information.
The system extracts symptom mentions from the input, maps them to LingShu concept entities, and invokes graph-based reasoning tools to infer potentially related syndromes, herbs, prescriptions, and acupoints.
The output includes structured graph evidence and decision-support suggestions, which are intended for reference and research exploration rather than direct clinical diagnosis or prescription.

Research Mode is designed for mechanism analysis and hypothesis generation.
It integrates entity search, relation type discovery, graph querying, and reasoning tools.
This mode supports candidate target prioritization for herbs, syndromes, symptoms, acupoints, and meridians, as well as association-oriented reasoning based on symptoms or diseases.
Unlike direct retrieval-oriented modes, Research Mode expands the reasoning process by default, making it suitable for exploring potential mechanisms, candidate targets, and evidence chains that can support biomedical and TCM research hypotheses.

\subsubsection{Knowledge Exploration and Visualization}

The platform also provides data browsing and graph visualization functions for interactive exploration of LingShu.
Users can search for a concept by name, inspect its normalized concept identifier, entity type, synonyms, and associated source evidence, and then explore its local neighborhood in the graph.
The visualization module supports interactive subgraph exploration, allowing users to trace multi-hop paths such as herb--chemical--gene associations, prescription--herb--therapeutic-efficacy links, syndrome--symptom--disease connections, and acupoint--therapeutic-efficacy--symptom relations.
For contextual quadruples, the platform preserves the contextual entity together with the head--tail relation.
This allows users to distinguish general binary associations from context-specific relations, such as syndrome-conditioned treatment, therapeutic-efficacy-mediated treatment, disease-conditioned drug effects, population-specific co-occurrence, prescription treatment through a constituent herb, and prescription modification through herb addition or removal.
Such visualization helps users inspect not only whether two entities are related, but also the clinical, mechanistic, compositional, or population context associated with the relation.

\subsubsection{Continuous Knowledge Curation}

To support continuous evolution of LingShu, we developed a human-in-the-loop data management and review platform for authorized knowledge curation.
The platform supports task assignment, entity-type-specific review, data validation, invalid relation inspection, metadata management, logical inference management, and data maintenance.
Trained curators can claim curation tasks according to entity type, inspect extracted terms together with their contextual evidence, update curation status, and flag uncertain records for further inspection.
In practice, trained curators continuously participate in the curation process, allowing newly extracted entities and relations to be progressively verified and incorporated into LingShu.
Before new knowledge is committed to the graph, schema validation is applied to ensure that submitted entities, triples, and contextual quadruples conform to predefined ontology constraints.
When multiple sources support the same concept-level relation, LingShu preserves separate evidence records while maintaining a unified relation representation in the conceptual graph.
This design enables LingShu to evolve continuously while preserving provenance, consistency, traceability, and human-auditable quality control.

\section{Discussion}

The construction of LingShu establishes a framework for integrating TCM and modern biomedicine within a unified, symptom-centric knowledge graph.
By systematically combining heterogeneous sources, including clinical electronic medical records, classical TCM texts, modern TCM books, and biomedical databases, LingShu integrates over 17.33 million atom-level entity records, 17.19 million semantic triples, and 22.29 million contextualized quadruples.
Unlike conventional biomedical knowledge graphs that are predominantly disease-centered, LingShu prioritizes symptom phenotypes as primary entities and introduces contextualized quadruples $(h, r, t, c)$ to capture conditional medical knowledge.

A central design principle of LingShu is its symptom-centric architecture.
In typical biomedical databases, symptoms are treated as attributes of diseases, limiting cross-disease phenotypic analysis and mechanistic exploration.
LingShu elevates symptoms to first-class entities, organizing the graph around symptom phenotypes.
This aligns with TCM diagnostic logic, where symptom patterns drive syndrome differentiation, while also supporting modern biomedical research that connects phenotypic manifestations to molecular mechanisms \cite{kohler2021human,putman2024monarch}.
Through this design, LingShu enables bidirectional mapping between clinical phenotypes and molecular targets, providing a shared layer for cross-paradigm knowledge alignment.

\begin{figure}[!tb]
    \centering
    \includegraphics[width=0.9\textwidth]{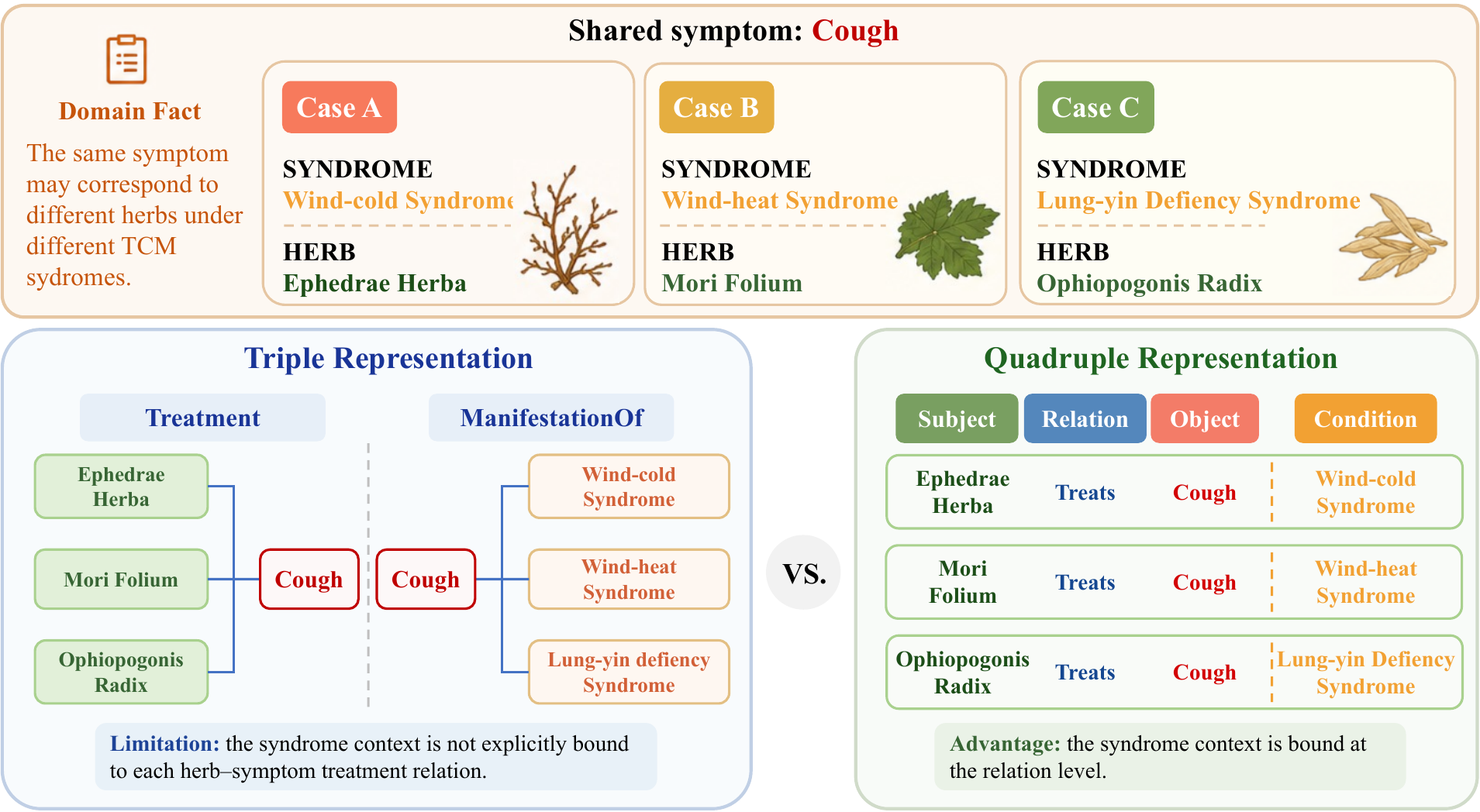}
    \caption{Triple and quadruple representations of syndrome-dependent treatment. For the shared symptom cough, triples separately encode herb--symptom treatment and symptom--syndrome associations, whereas contextualized quadruples bind each treatment relation to its syndrome context.}
    \label{fig:triple_vs_quad}
  \end{figure}

The introduction of contextualized quadruples is a key methodological innovation.
Medical statements are often conditional: therapeutic effects in TCM depend on specific syndromes, mechanism-based relations rely on molecular targets, and population-level associations hold only for particular cohorts.
Traditional triples $(h, r, t)$ cannot encode such conditionality without losing essential context.
Figure \ref{fig:triple_vs_quad} illustrates this limitation using syndrome-dependent treatment for a shared symptom.
In a triple representation, multiple herbs can be linked to the same symptom, and the symptom can also be linked to multiple syndromes, but the graph does not specify which syndrome constrains each herb--symptom treatment relation.
By extending triples to $(h, r, t, c)$, LingShu explicitly binds the syndrome context to the treatment relation itself.
Conventional triples are maintained for global connectivity, while quadruples handle conditional knowledge, including syndrome-dependent TCM treatments, mechanism-mediated drug--gene interactions, and population-specific associations derived from clinical EMRs.
This design improves semantic fidelity relative to standard triple-based representations \cite{guha1992contexts,mccarthy1997formalizing}.

LingShu further incorporates human-in-the-loop verification and quality control mechanisms.
Entities and relations extracted at the book scale or from EMRs are subjected to vector-based pre-screening, concept normalization, and type verification, supported by deep learning models embedded in our annotation system.
Population-specific co-occurrence relations were statistically validated, providing source-grounded evidence for cohort-specific symptom--herb and symptom--disease associations.
These verification processes enhance data reliability and traceability, while providing a stronger evidential basis for downstream exploration.

The contextualized knowledge graph also supports practical reasoning and interactive knowledge exploration.
By integrating graph-based reasoning with a RAG interface, LingShu facilitates efficient knowledge retrieval, candidate prioritization, and interpretable inference.
Global graph ranking and path-based semantic reasoning enable the system to identify candidate entities while exposing logical pathways connecting symptoms, interventions, syndromes, prescriptions, acupoints, and molecular targets.
On this basis, the web platform provides multiple Q\&A modes for direct retrieval, agent-based multi-step reasoning, free-text clinical exploration, and mechanism-oriented research, together with structured reasoning modes for symptom-driven, disease-driven, syndrome--prescription--herb, and acupuncture-related tasks.
In addition, the backend audit platform supports continuous quality control through updates from trained curators, enabling traceable knowledge curation.
Because LingShu preserves atom-level provenance, concept-level normalization, and human-reviewed curation records, it can be continuously updated as new clinical records, TCM texts, biomedical databases, and literature evidence become available.
Together, these functions transform LingShu from a static knowledge repository into an evidence-grounded reasoning and curation platform for decision-support exploration, integrative medicine research, and mechanism-based hypothesis generation.

Beyond methodological innovation, LingShu provides a computational framework for integrative medicine research.
By linking TCM entities such as symptoms, syndromes, herbs, prescriptions, meridians, and acupoints with modern biomedical entities including diseases, drugs, chemicals, genes, and molecular mechanisms, the graph enables cross-paradigm knowledge exploration.
Recent clinical and pharmacological studies have increasingly demonstrated the value of integrating traditional herbal therapies with modern biomedical research frameworks \cite{cheang2024traditional,yang2023traditional}.
Through these cross-level connections, LingShu allows researchers to explore potential molecular mechanisms underlying traditional therapeutic knowledge and to generate biologically testable hypotheses from empirical clinical observations.

Despite these strengths, limitations remain.
Current population-level associations are derived from a limited set of clinical cohorts, including COVID-19 and cirrhosis, and relation strengths are not yet quantitatively weighted.
Expanding patient cohorts and introducing probabilistic or quantitative relation weights would improve the representativeness and precision of inference.
In addition, although ontology-guided verification and human review improve data quality, large-scale knowledge acquisition from classical texts and heterogeneous medical documents may still be affected by OCR errors, ambiguous expressions, and incomplete context.
Future work will therefore require broader expert evaluation, richer benchmark datasets, and continuous refinement of extraction and verification procedures.
Furthermore, integrating multi-modal biomedical data, including omics, imaging, and real-time physiological measurements, would enrich the graph's contextual dimensions and enhance patient-specific modeling.

Overall, LingShu provides a scalable and interpretable infrastructure for bridging TCM and modern biomedicine through symptom-centric organization, contextualized knowledge representation, and evidence-grounded reasoning.

\section{Conclusion}

In this study, we presented LingShu, a large-scale contextualized knowledge graph that integrates 17.33 million atom-level entity records into a unified symptom-centric representation linking TCM and modern biomedicine. 
By combining classical literature, clinical electronic medical records, and biomedical resources with a hyper-relational quadruple structure, LingShu moves beyond conventional binary graph representations by explicitly encoding the contextual conditions associated with medical relations. 
This design strengthens the representation of syndrome-dependent, population-specific, and mechanism-related knowledge, while preserving conventional triples for broad semantic connectivity. 
The graph-based reasoning and RAG-supported web service further demonstrate how LingShu can translate contextualized knowledge into traceable exploration workflows for association analysis, decision-support exploration, and mechanism-oriented hypothesis generation.
Together, LingShu provides a scalable and interpretable infrastructure for AI-assisted integrative medicine research and future evidence-grounded decision-support studies.

\section*{Data Quality}

LingShu stores atom-level provenance for entities and relations. Records integrated from public databases retain the source database name and accession identifier, enabling traceability to the original resource. For knowledge extracted from texts, candidate entities, triples, and contextualized quadruples undergo rule-based schema validation and LLM-assisted evidence verification, in which the LLM checks whether each extraction is supported by its source text snippet. Given the large scale and continuous growth of the graph, trained domain experts perform daily incremental review and correction through the LingShu curation platform, providing ongoing quality control.

\section*{Data Availability}

The LingShu web platform is available for online browsing and interactive exploration at \url{http://www.tcmkg.com/}.
The dataset, including the ontology schema, entity records, semantic triples, contextualized quadruples, and supporting metadata, will be made publicly available for download upon acceptance of the manuscript.


\bibliographystyle{unsrt}
\bibliography{references}

\begin{thebibliography}{10}

\bibitem{fensel2020introduction}
Dieter Fensel, Umutcan {\c{S}}im{\c{s}}ek, Kevin Angele, Elwin Huaman, Elias
  K{\"a}rle, Oleksandra Panasiuk, Ioan Toma, J{\"u}rgen Umbrich, Alexander
  Wahler, Dieter Fensel, et~al.
\newblock Introduction: What is a knowledge graph?
\newblock {\em Knowledge Graphs: Methodology, Tools and Selected Use Cases},
  pages 1--10, 2020.

\bibitem{yu2017knowledge}
Tong Yu, Jinghua Li, Qi~Yu, Ye~Tian, Xiaofeng Shun, Lili Xu, Ling Zhu, and
  Hongjie Gao.
\newblock Knowledge graph for tcm health preservation: Design, construction,
  and applications.
\newblock {\em Artificial Intelligence in Medicine}, 77:48--52, 2017.

\bibitem{zheng2020tcmkg}
Ziqiang Zheng, Yongguo Liu, Yun Zhang, and Chuanbiao Wen.
\newblock Tcmkg: A deep learning based traditional chinese medicine knowledge
  graph platform.
\newblock In {\em 2020 IEEE International Conference on Knowledge Graph}, pages
  560--564. IEEE, 2020.

\bibitem{uniprot2019uniprot}
UniProt Consortium.
\newblock Uniprot: A worldwide hub of protein knowledge.
\newblock {\em Nucleic Acids Research}, 47(D1):D506--D515, 2019.

\bibitem{gene2019gene}
Gene~Ontology Consortium.
\newblock The gene ontology resource: 20 years and still going strong.
\newblock {\em Nucleic Acids Research}, 47(D1):D330--D338, 2019.

\bibitem{wu2019symmap}
Yang Wu, Feilong Zhang, Kuo Yang, Shuangsang Fang, Dechao Bu, Hui Li, Liang
  Sun, Hairuo Hu, Kuo Gao, Wei Wang, et~al.
\newblock Symmap: An integrative database of traditional chinese medicine
  enhanced by symptom mapping.
\newblock {\em Nucleic Acids Research}, 47(D1):D1110--D1117, 2019.

\bibitem{ru2014tcmsp}
Jinlong Ru, Peng Li, Jinan Wang, Wei Zhou, Bohui Li, Chao Huang, Pidong Li,
  Zihu Guo, Weiyang Tao, Yinfeng Yang, et~al.
\newblock Tcmsp: A database of systems pharmacology for drug discovery from
  herbal medicines.
\newblock {\em Journal of Cheminformatics}, 6:1--6, 2014.

\bibitem{bodenreider2004unified}
Olivier Bodenreider.
\newblock The unified medical language system: Integrating biomedical
  terminology.
\newblock {\em Nucleic Acids Research}, 32(suppl\_1):D267--D270, 2004.

\bibitem{donnelly2006snomed}
Kevin Donnelly et~al.
\newblock Snomed-ct: The advanced terminology and coding system for ehealth.
\newblock {\em Studies in Health Technology and Informatics}, 121:279, 2006.

\bibitem{schulz2023snomed}
Stefan Schulz, James~T Case, Peter Hendler, Daniel Karlsson, Michael Lawley,
  Ronald Cornet, Robert Hausam, Harold Solbrig, Karim Nashar, Catalina
  Mart{\'\i}nez-Costa, et~al.
\newblock Snomed ct and basic formal ontology--convergence or contradiction
  between standards? the case of “clinical finding”.
\newblock {\em Applied Ontology}, 18(3):207--237, 2023.

\bibitem{li2020real}
Linfeng Li, Peng Wang, Jun Yan, Yao Wang, Simin Li, Jinpeng Jiang, Zhe Sun,
  Buzhou Tang, Tsung-Hui Chang, Shenghui Wang, et~al.
\newblock Real-world data medical knowledge graph: Construction and
  applications.
\newblock {\em Artificial Intelligence in Medicine}, 103:101817, 2020.

\bibitem{gong2021smr}
Fan Gong, Meng Wang, Haofen Wang, Sen Wang, and Mengyue Liu.
\newblock Smr: Medical knowledge graph embedding for safe medicine
  recommendation.
\newblock {\em Big Data Research}, 23:100174, 2021.

\bibitem{ashley2016towards}
Euan~A Ashley.
\newblock Towards precision medicine.
\newblock {\em Nature Reviews Genetics}, 17(9):507--522, 2016.

\bibitem{packer2024traditional}
Milton Packer.
\newblock Traditional chinese medicine: cardiovascular drug development through
  a holistic framework, 2024.

\bibitem{wang2025tcmeval}
Zhe Wang, Meng Hao, Suyuan Peng, Yuyan Huang, Yiwei Lu, Keyu Yao, Xiaolin Yang,
  and Yan Zhu.
\newblock Tcmeval-sdt: a benchmark dataset for syndrome differentiation thought
  of traditional chinese medicine.
\newblock {\em Scientific Data}, 12(1):437, 2025.

\bibitem{wang2025rigorous}
Zeyu Wang, Dirui Xu, and Xin Chen.
\newblock Rigorous clinical trials for traditional chinese herbal medicine:
  Challenge and opportunity, 2025.

\bibitem{lowe2024persistent}
Bernd L{\"o}we, Anne Toussaint, Judith~GM Rosmalen, Wei-Lieh Huang, Christopher
  Burton, Angelika Weigel, James~L Levenson, and Peter Henningsen.
\newblock Persistent physical symptoms: definition, genesis, and management.
\newblock {\em The Lancet}, 403(10444):2649--2662, 2024.

\bibitem{subramanian2022symptoms}
Anuradhaa Subramanian, Krishnarajah Nirantharakumar, Sarah Hughes, Puja Myles,
  Tim Williams, Krishna~M Gokhale, Tom Taverner, Joht~Singh Chandan, Kirsty
  Brown, Nikita Simms-Williams, et~al.
\newblock Symptoms and risk factors for long covid in non-hospitalized adults.
\newblock {\em Nature medicine}, 28(8):1706--1714, 2022.

\bibitem{shin2023associations}
Hyewon Shin, William~N Dudley, Nickhill Bhakta, Madeline~R Horan, Zhaoming
  Wang, T~Robin Bartlett, Deokumar Srivastava, Yutaka Yasui, Justin~N Baker,
  Leslie~L Robison, et~al.
\newblock Associations of symptom clusters and health outcomes in adult
  survivors of childhood cancer: a report from the st jude lifetime cohort
  study.
\newblock {\em Journal of Clinical Oncology}, 41(3):497--507, 2023.

\bibitem{cheang2024traditional}
Iokfai Cheang, Wenming Yao, Yanli Zhou, Xu~Zhu, Gehui Ni, Xinyi Lu, Shengen
  Liao, Rongrong Gao, Fang Zhou, Jiangang Shen, et~al.
\newblock The traditional chinese medicine qiliqiangxin in heart failure with
  reduced ejection fraction: a randomized, double-blind, placebo-controlled
  trial.
\newblock {\em Nature Medicine}, 30(8):2295--2302, 2024.

\bibitem{yang2023traditional}
Yuejin Yang, Xiangdong Li, Guihao Chen, Ying Xian, Haitao Zhang, Yuan Wu,
  Yanmin Yang, Jianhua Wu, Chuntong Wang, Shenghu He, et~al.
\newblock Traditional chinese medicine compound (tongxinluo) and clinical
  outcomes of patients with acute myocardial infarction: the cts-ami randomized
  clinical trial.
\newblock {\em Jama}, 330(16):1534--1545, 2023.

\bibitem{ji2024jinlida}
Hangyu Ji, Xuefei Zhao, Xinyan Chen, Hui Fang, Huailin Gao, Geng Wei, Min
  Zhang, Hongyu Kuang, Baijing Yang, Xiaojun Cai, et~al.
\newblock Jinlida for diabetes prevention in impaired glucose tolerance and
  multiple metabolic abnormalities: the focus randomized clinical trial.
\newblock {\em JAMA internal medicine}, 184(7):727--735, 2024.

\bibitem{chandak2023building}
Payal Chandak, Kexin Huang, and Marinka Zitnik.
\newblock Building a knowledge graph to enable precision medicine.
\newblock {\em Scientific Data}, 10(1):67, 2023.

\bibitem{bang2023biomedical}
Dongmin Bang, Sangsoo Lim, Sangseon Lee, and Sun Kim.
\newblock Biomedical knowledge graph learning for drug repurposing by extending
  guilt-by-association to multiple layers.
\newblock {\em Nature Communications}, 14(1):3570, 2023.

\bibitem{lu2023sympgan}
Kezhi Lu, Kuo Yang, Hailong Sun, Qian Zhang, Qiguang Zheng, Kuan Xu, Jianxin
  Chen, and Xuezhong Zhou.
\newblock Sympgan: a systematic knowledge integration system for symptom--gene
  associations network.
\newblock {\em Knowledge-Based Systems}, 276:110752, 2023.

\bibitem{guha1992contexts}
Ramanathan~V Guha.
\newblock {\em Contexts: A formalization and some applications}.
\newblock Stanford University, 1992.

\bibitem{mccarthy1997formalizing}
John McCarthy, Sasa Buvac, et~al.
\newblock Formalizing context.
\newblock {\em Computing Natural Language, Stanford University}, pages 13--50,
  1997.

\bibitem{polanyi1958personal}
Michael Polanyi.
\newblock {\em Personal Knowledge: Towards a Post-Critical Philosophy}.
\newblock University of Chicago Press, Chicago, 1958.

\bibitem{mccarthy1994formalizing}
John McCarthy and Sasa Buvac.
\newblock Formalizing context.
\newblock In {\em AAAI Fall Symposium on Context in Knowledge Representation},
  pages 99--135. Citeseer, 1994.

\bibitem{lauschke2024pharmacogenomics}
Volker~M Lauschke, Yitian Zhou, and Magnus Ingelman-Sundberg.
\newblock Pharmacogenomics beyond single common genetic variants: the way
  forward.
\newblock {\em Annual Review of Pharmacology and Toxicology}, 64(1):33--51,
  2024.

\bibitem{iglesias2023comparison}
Ana Iglesias-Molina, Kian Ahrabian, Filip Ilievski, Jay Pujara, and Oscar
  Corcho.
\newblock Comparison of knowledge graph representations for consumer scenarios.
\newblock In {\em International Semantic Web Conference}, pages 271--289.
  Springer, 2023.

\bibitem{vrandevcic2014wikidata}
Denny Vrande{\v{c}}i{\'c} and Markus Kr{\"o}tzsch.
\newblock Wikidata: a free collaborative knowledgebase.
\newblock {\em Communications of the ACM}, 57(10):78--85, 2014.

\bibitem{lipscomb2000medical}
Carolyn~E Lipscomb.
\newblock Medical subject headings.
\newblock {\em Bulletin of the Medical Library Association}, 88(3):265, 2000.

\bibitem{canese2013pubmed}
Kathi Canese and Sarah Weis.
\newblock Pubmed: the bibliographic database.
\newblock {\em The NCBI Handbook}, 2(1), 2013.

\bibitem{noy2009bioportal}
Natalya~F Noy, Nigam~H Shah, Patricia~L Whetzel, Benjamin Dai, Michael Dorf,
  Nicholas Griffith, Clement Jonquet, Daniel~L Rubin, Margaret-Anne Storey,
  Christopher~G Chute, et~al.
\newblock Bioportal: Ontologies and integrated data resources at the click of a
  mouse.
\newblock {\em Nucleic Acids Research}, 37(suppl\_2):W170--W173, 2009.

\bibitem{kohler2021human}
Sebastian K{\"o}hler, Michael Gargano, Nicolas Matentzoglu, Leigh~C Carmody,
  David Lewis-Smith, Nicole~A Vasilevsky, Daniel Danis, Ganna Balagura, Gareth
  Baynam, Amy~M Brower, et~al.
\newblock The human phenotype ontology in 2021.
\newblock {\em Nucleic Acids Research}, 49(D1):D1207--D1217, 2021.

\bibitem{rappaport2017malacards}
Noa Rappaport, Michal Twik, Inbar Plaschkes, Ron Nudel, Tsippi Iny~Stein, Jacob
  Levitt, Moran Gershoni, C~Paul Morrey, Marilyn Safran, and Doron Lancet.
\newblock Malacards: An amalgamated human disease compendium with diverse
  clinical and genetic annotation and structured search.
\newblock {\em Nucleic Acids Research}, 45(D1):D877--D887, 2017.

\bibitem{weinreich2008orphanet}
Steffanie~S Weinreich, R~Mangon, JJ~Sikkens, ME~En Teeuw, and MC~Cornel.
\newblock Orphanet: A european database for rare diseases.
\newblock {\em Nederlands Tijdschrift Voor Geneeskunde}, 152(9):518--519, 2008.

\bibitem{harrison2021icd}
James~E Harrison, Stefanie Weber, Robert Jakob, and Christopher~G Chute.
\newblock Icd-11: An classification of diseases for the twenty-first century.
\newblock {\em BioMed Central Informatics and Decision Making}, 21:1--10, 2021.

\bibitem{himmelstein2017systematic}
Daniel~Scott Himmelstein, Antoine Lizee, Christine Hessler, Leo Brueggeman,
  Sabrina~L Chen, Dexter Hadley, Ari Green, Pouya Khankhanian, and Sergio~E
  Baranzini.
\newblock Systematic integration of biomedical knowledge prioritizes drugs for
  repurposing.
\newblock {\em elife}, 6:e26726, 2017.

\bibitem{morris2023scalable}
John~H Morris, Karthik Soman, Rabia~E Akbas, Xiaoyuan Zhou, Brett Smith,
  Elaine~C Meng, Conrad~C Huang, Gabriel Cerono, Gundolf Schenk, Angela
  Rizk-Jackson, et~al.
\newblock The scalable precision medicine open knowledge engine (spoke): a
  massive knowledge graph of biomedical information.
\newblock {\em Bioinformatics}, 39(2):btad080, 2023.

\bibitem{santos2022knowledge}
Alberto Santos, Ana~R Cola{\c{c}}o, Annelaura~B Nielsen, Lili Niu, Maximilian
  Strauss, Philipp~E Geyer, Fabian Coscia, Nicolai J~Wewer Albrechtsen, Filip
  Mundt, Lars~Juhl Jensen, et~al.
\newblock A knowledge graph to interpret clinical proteomics data.
\newblock {\em Nature biotechnology}, 40(5):692--702, 2022.

\bibitem{wishart2018drugbank}
David~S Wishart, Yannick~D Feunang, An~C Guo, Elvis~J Lo, Ana Marcu, Jason~R
  Grant, Tanvir Sajed, Daniel Johnson, Carin Li, Zinat Sayeeda, et~al.
\newblock Drugbank 5.0: A major update to the drugbank database for 2018.
\newblock {\em Nucleic Acids Research}, 46(D1):D1074--D1082, 2018.

\bibitem{kuhn2016sider}
Michael Kuhn, Ivica Letunic, Lars~Juhl Jensen, and Peer Bork.
\newblock The sider database of drugs and side effects.
\newblock {\em Nucleic Acids Research}, 44(D1):D1075--D1079, 2016.

\bibitem{kuhn2014stitch}
Michael Kuhn, Damian Szklarczyk, Sune Pletscher-Frankild, Thomas~H Blicher,
  Christian Von~Mering, Lars~J Jensen, and Peer Bork.
\newblock Stitch 4: Integration of protein--chemical interactions with user
  data.
\newblock {\em Nucleic Acids Research}, 42(D1):D401--D407, 2014.

\bibitem{szklarczyk2019string}
Damian Szklarczyk, Annika~L Gable, David Lyon, Alexander Junge, Stefan Wyder,
  Jaime Huerta-Cepas, Milan Simonovic, Nadezhda~T Doncheva, John~H Morris, Peer
  Bork, et~al.
\newblock String v11: Protein--protein association networks with increased
  coverage, supporting functional discovery in genome-wide experimental
  datasets.
\newblock {\em Nucleic Acids Research}, 47(D1):D607--D613, 2019.

\bibitem{pinero2020disgenet}
Janet Pi{\~n}ero, Juan~Manuel Ram{\'\i}rez-Anguita, Josep Sa{\"u}ch-Pitarch,
  Francesco Ronzano, Emilio Centeno, Ferran Sanz, and Laura~I Furlong.
\newblock The disgenet knowledge platform for disease genomics: 2019 update.
\newblock {\em Nucleic Acids Research}, 48(D1):D845--D855, 2020.

\bibitem{zhou2004ontology}
Xuezhong Zhou, Zhaohui Wu, Aining Yin, Lancheng Wu, Weiyu Fan, and Ruen Zhang.
\newblock Ontology development for unified traditional chinese medical language
  system.
\newblock {\em Artificial Intelligence in Medicine}, 32(1):15--27, 2004.

\bibitem{shu2024ispo}
Zixin Shu, Rui Hua, Dengying Yan, Chenxia Lu, Meng Ren, Hong Gao, Ning Xu, Jun
  Li, Hui Zhu, Jia Zhang, et~al.
\newblock Ispo: an integrated ontology of symptom phenotypes for semantic
  integration of traditional chinese medical data.
\newblock {\em Methods of Information in Medicine}, 63(05/06):164--175, 2024.

\bibitem{huang2018tcmid}
Lin Huang, Duoli Xie, Yiran Yu, Huanlong Liu, Yan Shi, Tieliu Shi, and
  Chengping Wen.
\newblock Tcmid 2.0: a comprehensive resource for tcm.
\newblock {\em Nucleic acids research}, 46(D1):D1117--D1120, 2018.

\bibitem{fang2021herb}
ShuangSang Fang, Lei Dong, Liu Liu, JinCheng Guo, LianHe Zhao, JiaYuan Zhang,
  DeChao Bu, XinKui Liu, PeiPei Huo, WanChen Cao, et~al.
\newblock Herb: A high-throughput experiment-and reference-guided database of
  traditional chinese medicine.
\newblock {\em Nucleic Acids Research}, 49(D1):D1197--D1206, 2021.

\bibitem{liu2020tcmio}
Zhihong Liu, Chuipu Cai, Jiewen Du, Bingdong Liu, Lu~Cui, Xiude Fan, Qihui Wu,
  Jiansong Fang, and Liwei Xie.
\newblock Tcmio: a comprehensive database of traditional chinese medicine on
  immuno-oncology.
\newblock {\em Frontiers in pharmacology}, 11:439, 2020.

\bibitem{ye2010hit}
Hao Ye, Li~Ye, Hong Kang, Duanfeng Zhang, Lin Tao, Kailin Tang, Xueping Liu,
  Ruixin Zhu, Qi~Liu, YZ~Chen, et~al.
\newblock Hit: linking herbal active ingredients to targets.
\newblock {\em Nucleic acids research}, 39(suppl\_1):D1055--D1059, 2010.

\bibitem{yan2022hit}
Deyu Yan, Genhui Zheng, Caicui Wang, Zikun Chen, Tiantian Mao, Jian Gao,
  Yu~Yan, Xiangyi Chen, Xuejie Ji, Jinyu Yu, et~al.
\newblock Hit 2.0: an enhanced platform for herbal ingredients' targets.
\newblock {\em Nucleic acids research}, 50(D1):D1238--D1243, 2022.

\bibitem{zhang2022sofda}
Yanqiong Zhang, Ning Wang, Xia Du, Tong Chen, Zecong Yu, Yuewen Qin, Wenjia
  Chen, Meng Yu, Ping Wang, Huamin Zhang, et~al.
\newblock Sofda: An integrated web platform from syndrome ontology to
  network-based evaluation of disease-syndrome-formula associations for
  precision medicine.
\newblock {\em Science Bulletin}, 67(11):1097--1101, 2022.

\bibitem{zhou2019research}
Yang Zhou, Xingliang Qi, Yi~Huang, and Fangning Ju.
\newblock Research on construction and application of tcm knowledge graph based
  on ancient chinese texts.
\newblock In {\em IEEE/wic/acm International Conference on Web
  Intelligence-companion Volume}, pages 144--147, 2019.

\bibitem{zhang2024traditional}
Yichong Zhang and Yongtao Hao.
\newblock Traditional chinese medicine knowledge graph construction based on
  large language models.
\newblock {\em Electronics}, 13(7):1395, 2024.

\bibitem{duan2025research}
Yuchen Duan, Qingqing Zhou, Yu~Li, Chi Qin, Ziyang Wang, Hongxing Kan, and Jili
  Hu.
\newblock Research on a traditional chinese medicine case-based
  question-answering system integrating large language models and knowledge
  graphs.
\newblock {\em Frontiers in Medicine}, 11:1512329, 2025.

\bibitem{qu2024review}
Xiaolong Qu, Ziwei Tian, Jinman Cui, Ruowei Li, Dongmei Li, and Xiaoping Zhang.
\newblock A review of knowledge graph in traditional chinese medicine:
  analysis, construction, application and prospects.
\newblock {\em Computers, Materials, \& Continua}, 81(3):3583, 2024.

\bibitem{yan2025artificial}
Dengying Yan, Qiguang Zheng, Kai Chang, Rui Hua, Yiming Liu, Jingyan Xue, Zixin
  Shu, Yunhui Hu, Pengcheng Yang, Yu~Wei, et~al.
\newblock Artificial intelligence in traditional chinese medicine: from systems
  biological mechanism discovery, real-world clinical evidence inference to
  personalized clinical decision support.
\newblock {\em Chinese Journal of Natural Medicines}, 23(11):1310--1328, 2025.

\bibitem{alexander2006rdf}
Nicole Alexander and Siva Ravada.
\newblock Rdf object type and reification in the database.
\newblock In {\em 22nd International Conference on Data Engineering}, pages
  93--93. IEEE, 2006.

\bibitem{giunti2021representing}
Marco Giunti, Giuseppe Sergioli, Giuliano Vivanet, and Simone Pinna.
\newblock Representing n-ary relations in the semantic web.
\newblock {\em Logic Journal of the Igpl}, 29(4):697--717, 2021.

\bibitem{serafini2012contextualized}
Luciano Serafini and Martin Homola.
\newblock Contextualized knowledge repositories for the semantic web.
\newblock {\em Journal of Web Semantics}, 12:64--87, 2012.

\bibitem{jiang2019role}
Tianwen Jiang, Tong Zhao, Bing Qin, Ting Liu, Nitesh~V Chawla, and Meng Jiang.
\newblock The role of "condition" a novel scientific knowledge graph
  representation and construction model.
\newblock In {\em Proceedings of the 25th ACM SIGKDD International Conference
  on Knowledge Discovery \& Data Mining}, pages 1634--1642, 2019.

\bibitem{jiang2020biomedical}
Tianwen Jiang, Qingkai Zeng, Tong Zhao, Bing Qin, Ting Liu, Nitesh~V Chawla,
  and Meng Jiang.
\newblock Biomedical knowledge graphs construction from conditional statements.
\newblock {\em IEEE/ACM Transactions on Computational Biology and
  Bioinformatics}, 18(3):823--835, 2020.

\bibitem{xu2023conditional}
Zhangbiao Xu, Botao Zhang, Jinguang Gu, and Feng Gao.
\newblock Conditional knowledge extraction using contextual information
  enhancement.
\newblock {\em Applied Sciences}, 13(8):4954, 2023.

\bibitem{qwen3.5}
{Qwen Team}.
\newblock {Qwen3.5}: Towards native multimodal agents, February 2026.

\bibitem{zou2021phenonizer}
Qunsheng Zou, Kuo Yang, Kai Chang, Xiaoping Zhang, Xiaodong Li, and Xuezhong
  Zhou.
\newblock Phenonizer: A fine-grained phenotypic named entity recognizer for
  chinese clinical texts.
\newblock In {\em 2021 IEEE International Conference on Bioinformatics and
  Biomedicine}, pages 3963--3970. IEEE, 2021.

\bibitem{gan2023network}
Xiao Gan, Zixin Shu, Xinyan Wang, Dengying Yan, Jun Li, Shany Ofaim, R{\'e}ka
  Albert, Xiaodong Li, Baoyan Liu, Xuezhong Zhou, et~al.
\newblock Network medicine framework reveals generic herb-symptom effectiveness
  of traditional chinese medicine.
\newblock {\em Science advances}, 9(43):eadh0215, 2023.

\bibitem{yao2022react}
Shunyu Yao, Jeffrey Zhao, Dian Yu, Nan Du, Izhak Shafran, Karthik~R Narasimhan,
  and Yuan Cao.
\newblock React: Synergizing reasoning and acting in language models.
\newblock In {\em The eleventh international conference on learning
  representations}, 2022.

\bibitem{putman2024monarch}
Tim~E Putman, Kevin Schaper, Nicolas Matentzoglu, Vincent~P Rubinetti, Faisal~S
  Alquaddoomi, Corey Cox, J~Harry Caufield, Glass Elsarboukh, Sarah Gehrke,
  Harshad Hegde, et~al.
\newblock The monarch initiative in 2024: an analytic platform integrating
  phenotypes, genes and diseases across species.
\newblock {\em Nucleic acids research}, 52(D1):D938--D949, 2024.

\end{thebibliography}

\end{document}